\pdfoutput=1
\documentclass[preprint,authoryear]{elsarticle}
\usepackage[breaklinks=true]{hyperref}
\usepackage{inputenc}
\usepackage{csquotes}
\usepackage{epsfig}
\usepackage{amsmath}
\usepackage{amssymb}
\usepackage{pifont}
\usepackage{siunitx}
\usepackage{amsfonts}
\usepackage{booktabs}
\usepackage{multirow}
\usepackage{caption}
\usepackage{subcaption}
\usepackage{algorithm2e}
\usepackage{textcomp}
\usepackage{xspace}
\usepackage{multimedia}
\usepackage{graphicx}
\usepackage{float}
\usepackage{wrapfig}
\usepackage{pgfplots}
\pgfplotsset{compat=1.17}
\usetikzlibrary{spy}
\usepackage[toc,page]{appendix}
\usepackage[english]{babel}
\usepackage{overpic}
\usepackage{color,soul}
\usepackage{etoolbox}

\BeforeBeginEnvironment{wrapfigure}{\setlength{\intextsep}{0pt}}

\newcommand{\cmark}{\ding{51}}%
\newcommand{\xmark}{\ding{55}}%

\newcommand{\mabbr}{SOCRATES}
\newcommand{\mname}{\textbf{S}tere\textbf{O} \textbf{C}ame\textbf{RA} \textbf{T}rap for monitoring of biodiv\textbf{E}r\textbf{S}ity}

\newcommand{\plittersdorf}{\emph{Tierpark Plittersdorf}}

\makeatletter
\def\blfootnote{\gdef\@thefnmark{}\@footnotetext}
\makeatother

\begin{document}

\allowdisplaybreaks

\title{SOCRATES: A Stereo Camera Trap for Monitoring of Biodiversity}

\author[1]{Timm Haucke}
\ead{haucke@cs.uni-bonn.de}
\author[2]{Hjalmar S. Kühl}
\ead{kuehl@eva.mpg.de}
\author[1]{Volker Steinhage}
\ead{steinhage@cs.uni-bonn.de}

\date{\today}

\address[1]{University of Bonn, Institute of Computer Science IV, Friedrich-Hirzebruch-Allee 8, Bonn 53115, Germany}
\address[2]{German Centre for Integrative Biodiversity Research (iDiv) Halle-Jena-Leipzig, Puschstrasse 4, 04103 Leipzig, Germany}

\hyphenation{SO-CRA-TES}

\begin{abstract}
    \noindent
    The development and application of modern technology is an essential basis for the efficient monitoring of species in natural habitats to assess the change of ecosystems, species communities and populations, and in order to understand important drivers of change.
    For estimating wildlife abundance, camera trapping in combination with 3D (three-dimensional) measurements of habitats is highly valuable. Additionally, 3D information improves the accuracy of wildlife detection using camera trapping. 
    This study presents a novel approach to 3D camera trapping featuring highly optimized hardware and software. This approach employs stereo vision to infer 3D information of natural habitats and is designated as \mname{} (\mabbr{}). 
    A comprehensive evaluation of \mabbr{} shows not only a $3.23\%$ improvement in animal detection (bounding box $\text{mAP}_{75}$) but also its superior applicability for estimating animal abundance using camera trap distance sampling.
    The software and documentation of \mabbr{} is provided at \href{https://github.com/timmh/socrates}{https://github.com/timmh/socrates}.
\end{abstract}


\maketitle

\blfootnote{
\\
© 2022. This manuscript version is made available under the CC-BY-NC-ND 4.0 license \href{https://creativecommons.org/licenses/by-nc-nd/4.0/}{https://creativecommons.org/licenses/by-nc-nd/4.0/}
}

\clearpage


\section{Introduction}

\noindent 
The utilization of modern technology is commonplace in large-scale commercial, civil, and strategic projects. 
But terrestrial ecology, in particular, is less well equipped. The potential of modern sensors, remote sensing, automated laboratory procedures, and complex data processing is hardly utilized in this field. This absence is one important reason why there is no long-term large-scale automated monitoring of biodiversity (as established for climate research).

\subsection{AMMODs Framework}
\noindent 
To foster the adaption of modern technologies for the development of automated, reliable, and verifiable biodiversity monitoring, a network of \textbf{A}utomated \textbf{M}ultisensor stations for \textbf{Mo}nitoring of species \textbf{D}iversity (AMMODs) is proposed by \citet{waegele_ammod} to pave the way for a new generation of biodiversity assessment stations. The AMMODs network approach combines cutting-edge technologies with biodiversity informatics and expert systems to conserve expert knowledge. The sensors employed in AMMODs range from traps for DNA barcoding over camera traps for visual monitoring, bioacoustics, to plant volatile compound detectors. 
The general concept, the hardware and software components, and the setup of AMMODs are described by \citet{waegele_ammod}.
Within the AMMOD project, \mabbr{} is specifically concerned with the visual monitoring of wildlife. \mabbr{} derives and utilizes depth information as a third dimension in addition to the regular two dimensions (that is, the horizontal and vertical pixel coordinates) of conventional camera trap images. \footnote{In this contribution, the term \textit{depth} refers to the distance between the camera trap and the observed scene.}
\subsection{\mabbr{} Contributions}
%

\noindent $\bullet$
\textit{\textbf{Detection and localization of animals}} in camera trap images are often unreliable. The additional depth information provided by \mabbr{} fosters the accuracy and reliability of visual animal detection and animal localization within the monitored natural habitat.

\noindent $\bullet$
\textit{\textbf{Abundance estimation}}, using methods such as camera trap distance sampling (CTDS), is traditionally performed using a combination of commercial camera trap hardware and very laborious manual workflows. \mabbr{} instead provides depth information in a \textit{\textbf{fully automated}} way using stereo vision.

\noindent $\bullet$
\textit{\textbf{Reproducibility and accessibility}} for practitioners. This study takes the practitioner's perspective and provides detailed setup and operational instructions.

\subsection{Related Work}
Related work is reported using two perspectives. First, recent progress with respect to visual object detection in images and video clips is reported. Second, an overview on approaches to estimate the density and abundance of unmarked animal populations using camera traps is given.
\subsection{Visual Animal Detection}
State-of-the-art approaches to visual animal detection utilize 2D deep learning object detection methods. Given a 2D color or grayscale image, these methods learn to predict a set of bounding boxes, e.g. given by the 2D location of the upper-left and bottom-right corner of an axis-aligned rectangle fully enclosing the animal. Object detection methods might be extended to perform \emph{instance segmentation}, where not only bounding boxes are predicted, but whether any pixel in the image belongs to the respective object (binary masks). Deep learning object detection and instance segmentation models usually consist of two parts. The backbone takes in the original image and produces a hierarchy of \emph{feature maps} that encode higher- and higher-level information about the image. These feature maps are then used by the object detection or instance segmentation model to predict the bounding boxes and binary masks themselves. A general issue of deep learning models is that their training requires immense amounts of \emph{annotated} training data. Annotated data is raw data associated with corresponding labels, which can be of different modalities (e.g. object classes occurring in the image, bounding boxes around objects of interest, pixel-wise masks of such objects, etc.). These labels must often be created manually and are therefore costly to obtain. This requirement of large annotated training datasets is slightly relaxed by transfer learning. In transfer learning, the backbone is first pre-trained to perform some task involving a very large training dataset, e.g., performing image classification on ImageNet \citep{imagenet}. Visual concepts learned by such backbones have been shown to be generally useful and not just applicable to the pre-training task \citep{cnnvis}. The backbone is then fine-tuned on the target task, which usually involves a much smaller dataset.
%
%
\subsubsection{Abundance Estimation}
There exist a number of methods to estimate the density and abundance of unmarked animal populations using camera traps, e.g. the random encounter model (REM) \citep{rowcliffe2008estimating}, the random encounter and staying time model (REST) \citep{nakashima2018estimating}, the time-to-event model (TTE), space-to-event model (STE), instantaneous estimator (IS) \citep{moeller2018three} and camera trap distance sampling \citep{camera_trap_distance_sampling}. All of these require an estimation of the effective area surveyed by the camera trap. This area is not simply given by the optical constraints of the camera, instead it is influenced by factors such as environmental occlusion and the range of the passive infrared sensor which may not perform consistently at all locations within the camera's viewshed. The effective area surveyed is statistically inferred by using the distances of the observed animals. Although there are approaches that estimate these distances (semi) automatically \citep{haucke2022overcoming,johanns2022automated}, they either require laborious capture of reference material \citep{haucke2022overcoming} or might not generalize to extreme scenarios such as very close-up scenes within $\SI{3}{\meter}$ of the camera \citep{johanns2022automated,auda_johanns_issues}.

\section{Methods \& Data}
\subsection[The SOCRATES Stereovision Sensor Platform]{The \mabbr{} Stereovision Sensor Platform}

The \mabbr{} camera trap system comprises a cost and power efficient stereovision sensor platform as well as state-of-the-art animal detection software based on a deep learning software architecture. The experimental evaluation of \mabbr{} will utilize a representative RGB-D dataset generated by \mabbr{} in the wildlife park Plittersdorf located in Bonn, Germany, exhibiting fallow deer.

\begin{figure}
     \centering
     \begin{subfigure}[b]{0.49\textwidth}
         \centering
         \includegraphics[width=\textwidth]{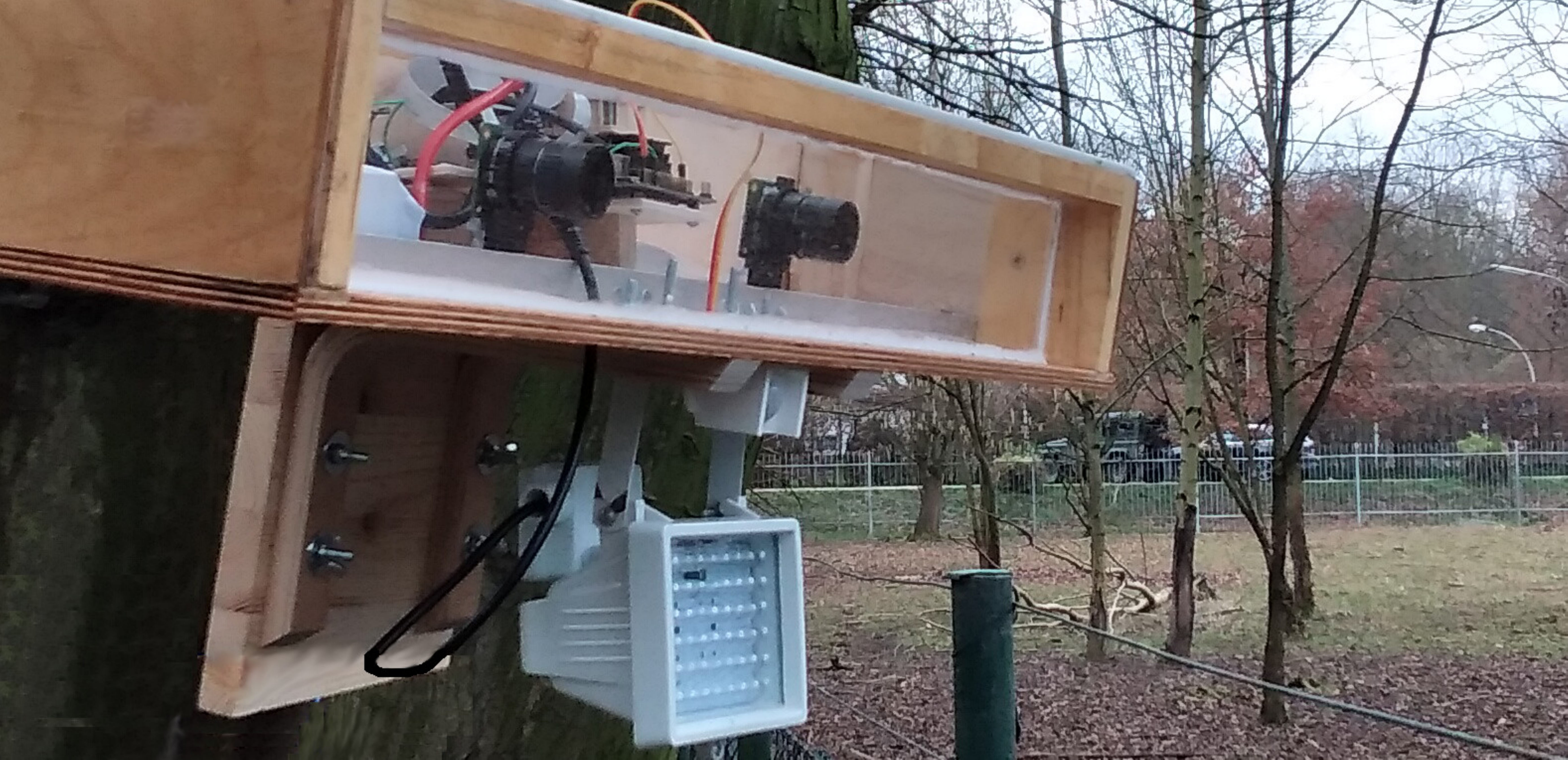}
     \end{subfigure}
     \hfill
     \begin{subfigure}[b]{0.49\textwidth}
         \centering
         \includegraphics[width=\textwidth]{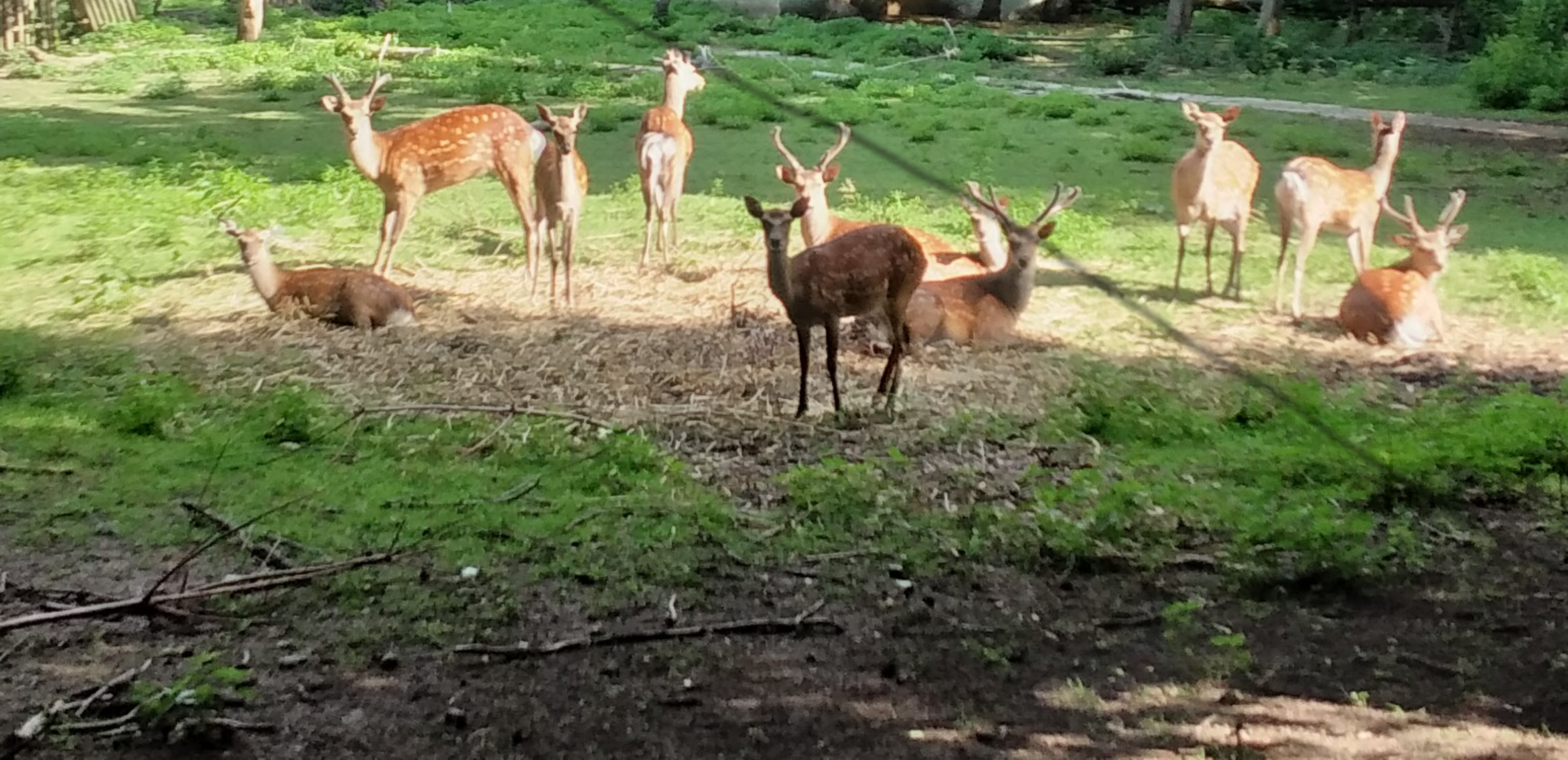}
     \end{subfigure}
    \caption{Introducing the \mabbr{} camera trap system located in the wildlife park Plittersdorf located in Bonn, Germany, showing European fallow deer (Dama dama) and Sika deer (Cervus nippon).}
    \label{fig:mount}
\end{figure}


The central goal of \mabbr{} is to infer depth information. Computer stereovision is the well-established approach to image-based depth estimation. By comparing information about an observed scene from two differing camera perspectives depth information can be derived. Mostly, both cameras in a stereo setup are displaced horizontally from one another yielding scene observations in terms of a left image and a right images. Computer stereo vision can be seen as the technical analogue to human stereopsis, that is, human perception of depth and three-dimensional structure by combining visual information from two eyes.


\subsubsection{Depth by Stereovision}\label{sec:stereo}
\vspace{1ex}

\noindent Depth is derived by stereo vision using two steps:
\begin{enumerate}
    \item \textit{Stereo Matching.} Given a stereo-pair image (i.e., left and right image) of an observed scene, a stereo matching model infers the so-called \textit{disparity map}. The disparity map $\mathbf{D}$ represents the position difference $\mathbf{d}$ of every observed scene point between the left and right image, when viewed in the right image. The disparity is inversely proportional to the distance from the stereo setup to the observed scene point depicted in the corresponding image points.
    \footnote{Readers unfamiliar with this topic might use their human stereopsis by looking at a near object and a distant object alternately with only their left and right eyes, respectively. The farther object will shift less than the nearer object.}
    \item \textit{Depth Computation.} Once the disparity map is derived, the disparity values $\mathbf{d}$ are used in combination with the so-called \textit{extrinsic parameters} of the stereovision setup to compute the absolute depth value for each observed scene point. The extrinsic parameters are (1) the baseline $b$ (physical distance between left and right cameras) and (2) the common focal length of both cameras $f$ (distance between the lens and camera sensor). The absolute depth value $\mathbf{z}$ for an observed scene point is then simply given by: $\mathbf{z} = \frac{b \cdot f}{\mathbf{d}}$. Thus, the lower the disparity, the larger the depth of the observed scene point.
\end{enumerate}

\begin{figure}
    \centering
    \includegraphics[width=\textwidth]{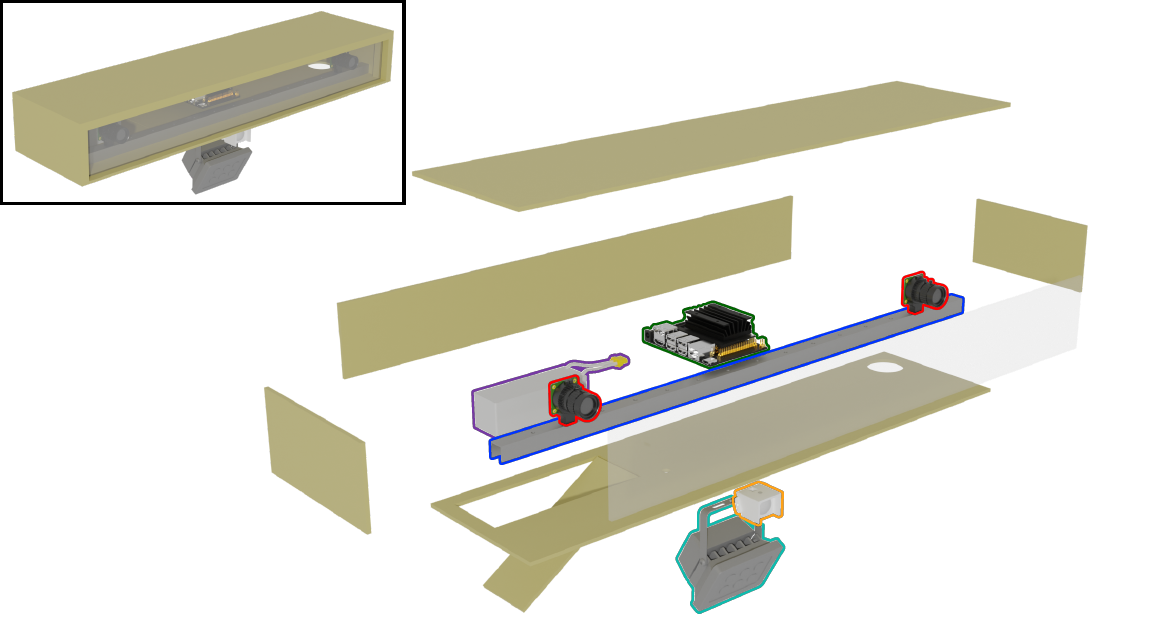}
    \caption{3D visualization of \mabbr{}. The important components will be covered in the following text and are here highlighted, i.e., cameras (red outline), baseline rail (blue outline), control unit (green outline), battery (violet outline), infrared illumination (turquoise outline), passive IR sensor (orange outline). Details such as power supply, wiring and screws are omitted. Some parts of the model are obtained from external sources\protect\footnotemark.}
    \label{fig:ct}
\end{figure}

\footnotetext{The following 3D parts of the model are obtained from external sources: Raspberry Pi HQ cameras \citep{3d_cam}, Jetson Nano Devkit \citep{3d_jetson}, Infrared illuminator \citep{3d_ir}, PIR sensor case \citep{3d_pir}, LiPo battery \citep{3d_bat}.}

\subsubsection[Hardware design of SOCRATES]{Hardware design of \mabbr{}}
\vspace{1ex}
\noindent The \mabbr{} stereovision platform is optimized for
\vspace{-1ex}
\begin{enumerate}
\item operability 
\begin{enumerate}
\item at day and night time as well as 
\item for a wide range of animal-camera distances, 
\end{enumerate}
\item effective and efficient power supply, 
\item hardware and construction costs, 
\item weather resistance.
\end{enumerate}
The following section covers in detail the technical implementation and how we address these design goals. We first address the stereo camera design (\textbf{cameras and Baseline}, design goals 1. and 3.). The raw data produced by the cameras is processed and stored by the \textbf{control} unit (design goals 2. and 3.). Weather-resistance (design goal 4.) is provided by the \textbf{case}. Infrared \textbf{motion Detection} and \textbf{illumination} facilitate energy efficiency (design goal 2.) and operability at night time (design goal 1. (a)). We additionally describe in detail the \textbf{power supply}, how we obtain animal-camera distances using \textbf{stereo correspondence} and how the captured data may be transferred using different \textbf{connectivity} options.
\newline\newline
\noindent \textbf{Cameras and Baseline:} A pair of Raspberry Pi High Quality Cameras are chosen for their cost-effectiveness and the high sensitivity of their Sony IMX477 sensor \citep{imx477_datasheet}. Interchangeable lenses allow adaptation to specific scenarios (i.e. shorter focal lengths for close-up scenes, higher focal lengths for more distant objects). Removal of the infrared filter allows sufficient exposure at night using artificial infrared illumination. The cameras have an additional Bayer filter above the sensor, which is usually responsible for filtering different wavelengths to create a color image. We leave this filter intact to not risk damaging the sensor itself. As near-infrared illumination (either from the environment or the illuminator) illuminates all color bands, we do not try to recover any color information and instead average all bands to obtain a grayscale image.

The cameras are mounted on a $\SI{77.5}{\centi\meter}$ long U-shaped aluminum rail with holes drilled at
regular intervals to allow configuration of different baseline distances between both cameras. Both cameras are connected through $\SI{50}{\centi\meter}$ long ribbon cables to the two MIPI CSI-2 interfaces of an NVIDIA Jetson Nano Developer Kit.

Both design aspects, i.e., the interchangeable high quality lenses as well as the configurable baseline construction, allow for adaptation to specific scenarios, e.g., free fields, feeding places, animal crosses, green bridges, etc. where animals are observable at different distances.

\noindent \textbf{Control:} We use an NVIDIA Jetson Nano Developer Kit as the central control and storage unit. It is responsible for taking motion detection signals from the PIR sensor, turning on the power to the IR illuminator, capturing, encoding, and archiving image material from the cameras. We decided on the Jetson Nano for the following reasons: (1) compared to most single-board computers, it provides two MIPI CSI-2 interfaces for the two cameras, (2) it provides a powerful GPU that can be used for encoding video efficiently, and (3) it supports a power-efficient hibernation mode (SC7) from which it can be activated through its general-purpose input/output (GPIO) pins by adjusting the Linux device tree accordingly. The raw RGB video material is encoded on the Jetson Nano's GPU by synchronizing the left and right image streams, concatenating them horizontally, and compressing the resulting video of resolution $2 \times 1920 \times 1080$ using the HEVC video codec \citep{hevc}. In our experiments, this results in a bitrate of roughly 6.7Mbit/s. The Jetson Nano uses a 128GB microSDXC card for persistent storage.



\noindent \textbf{Case:} To make \mabbr{} as weather-resistant as possible, most components are placed inside a single weather-proof case. The case is made of $\SI{0.8}{\centi\meter}$ thick birch plywood and is $\SI{80}{\centi\meter}$ wide, $\SI{11.6}{\centi\meter}$ high and $\SI{20}{\centi\meter}$ deep. We decided for a very wide case to be able to adapt the baseline of the stereo camera to different configurations. The front of the case is shielded by a piece of acrylic glass. In the bottom, we add a $\SI{4}{\centi\meter}$ diameter hole for ventilation, which is covered from the inside with an insect screen. The battery is mounted via Velcro strip onto a hatch in the bottom of the case, to allow quick replacement. We add two further holes for the wiring of the IR illuminator and motion detector, respectively, both of which are sealed using silicone. The top of the case is sealed using a silicone strip and secured by screws, which can be loosened to take it off for maintenance. All exposed wooden parts are further treated with marine varnish for weather resistance.

\noindent \textbf{Motion Detection:} Like most camera traps, we use a \textbf{p}yroelectric \textbf{i}nfra\textbf{r}ed (PIR) sensor for detecting motion and thereby triggering capture. We choose an HC-SR501 PIR sensor due to its compatibility with the $\SI{3.3}{\volt}$ GPIO pins of the Jetson Nano. We initially mounted the PIR sensor inside the case, just behind the acrylic glass. However, we found that this severely impaired the ability of the sensor to detect any kind of motion outside the case. This is because acrylic glass is opaque around wavelengths of $\SI{10}{\micro\meter}$ \citep{plexiglass_transmission}, which corresponds to the body temperatures of most animals. Therefore, we mount the PIR sensor in a separate 3D printed weatherproof casing below the main case.

\noindent \textbf{Illumination:} We employ a simple $\SI{12}{\watt}$, $\SI{850}{\nano\meter}$ infrared illuminator to ensure properly exposed images at night without disturbing most animals. The illuminator has a weatherproof case and is mounted on the bottom of the main case. The $\SI{12}{\volt}$ power supply is switched by a Jetson Nano GPIO pin using an IRLZ44NPBF MOSFET.

\noindent \textbf{Power supply:} All components are powered by a lithium ion polymer battery due to their high power density. We employ a battery with a theoretical capacity of $\SI{236.8}{\watt\hour}$ ($\SI{16000}{\milli\ampere\hour}$ at $\SI{14.8}{\volt}$). A generic 4S balancer circuit board provides over-discharge protection. The variable voltage of the battery is then regulated to $\SI{5}{\volt}$ for the Jetson Nano and $\SI{12}{\volt}$ for the infrared illuminator by Mean Well SCW20A-05 and SCW12A-12 converters, respectively.

\noindent \textbf{Stereo correspondence:} The central goal of \mabbr{} is to infer depth information through stereo vision. In the natural world, as well as in computer vision, this is achieved by solving the stereo correspondence problem. To solve the stereo correspondence problem efficiently, the left and right images must be \emph{rectified}. To obtain an accurate rectification, the intrinsic (internal camera parameters) and extrinsic (rotation and translation between the cameras) parameters have to be obtained by a calibration procedure. For the calibration of the intrinsic parameters, a calibration object (e.g. checkerboard pattern printed on cardboard) has to be captured by the camera(s) to be able to associate 3D points in the scene with 2D points in the resulting image. To obtain the extrinsic parameters, eight or more correspondences between images of points in the projections of both cameras must be established \citep{8pointalg}. We perform both intrinsic and extrinsic calibration using \textit{Kalibr} \citep{kalibr} with a grid of $4 \times 3$ AprilTags \citep{apriltag} mounted on a wooden board as the calibration target. During the setup of \mabbr{}, the calibration target is manually moved through the scene such that it covers as much as possible of each camera's field of view. After \mabbr{} is assembled and calibrated, calibration does not have to be repeated when deployed to different locations, as the calibration is not dependend not on a specific location but only on the camera configuration. Given the intrinsic and extrinsic parameters, we rectify the images of both cameras and compute the disparity of each pixel using \cite{crestereo}.

\noindent \textbf{Connectivity:} \mabbr{} may transmit the recorded data via three different means: wired ethernet cable, wireless LAN (Edimax EW-7811UN) or cellular connection (Huawei E3372H). If no basestation is available, we use the cellular connection to manually download the captured data. Otherwise, we connect via wireless LAN and the CoAP protocol \citep{coap} to the \emph{AMMOD Basestation} \citep{waegele_ammod,basestation}, which in turn uploads the captured data to the \emph{AMMOD Portal} (c.f. section \ref{sec:ammod_portal}).

\subsection{Data}

\begin{figure}
     \centering
     \begin{subfigure}[b]{\textwidth}
         \centering
         \includegraphics[width=\textwidth]{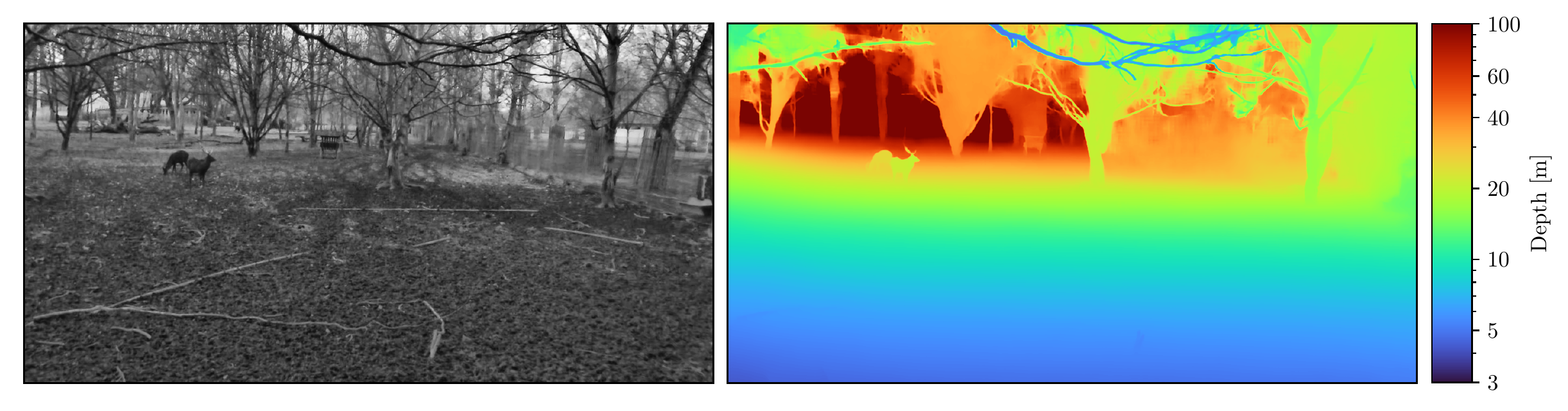}
     \end{subfigure}
     \hfill
     \begin{subfigure}[b]{\textwidth}
         \centering
         \includegraphics[width=\textwidth]{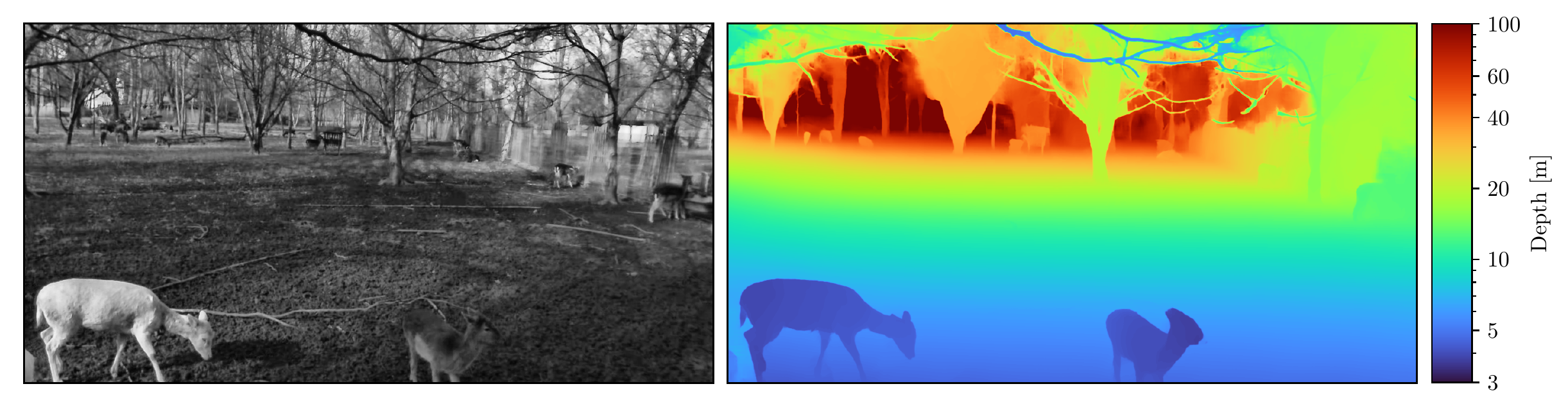}
     \end{subfigure}
    \caption{Samples of data collected in the \plittersdorf{}. The left shows the grayscale image of the left camera, the right image the colorcoded depth map obtained using stereo correspondence.}
    \label{fig:data_samples}
\end{figure}

\begin{figure}
     \centering
     \begin{subfigure}[b]{\textwidth}
         \centering
         \includegraphics[width=\textwidth]{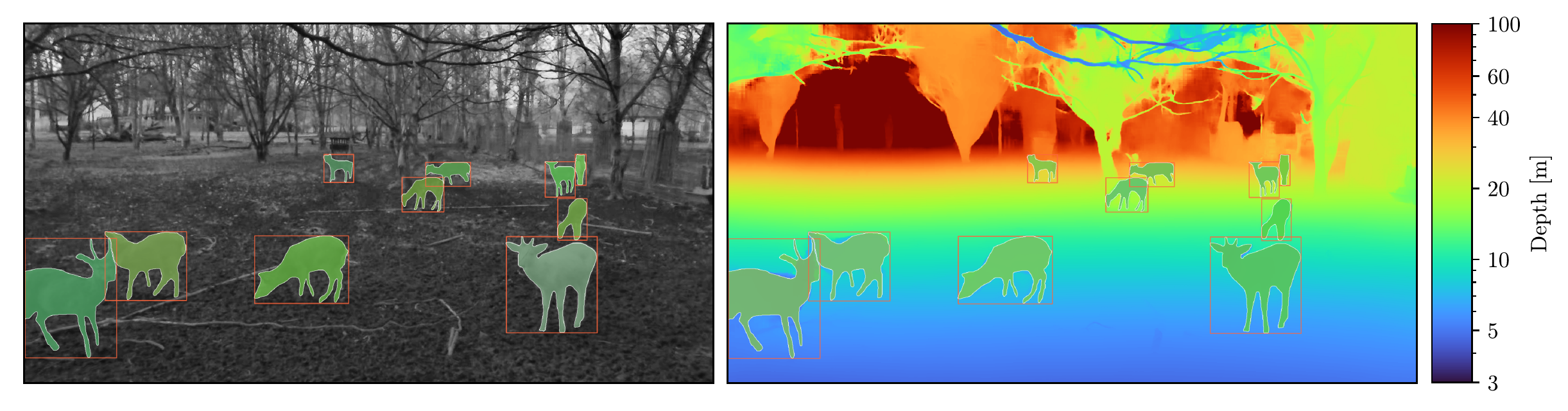}
     \end{subfigure}
    \caption{Examples of instance segmentation and bounding box annotations overlaid on top of the grayscale image of the left camera (left) and the colorcoded depth map (right).}
    \label{fig:labeling_samples}
\end{figure}

We deployed \mabbr{} in the \plittersdorf{} in order to evaluate the hardware and software. Details about this deployment may be found in section \ref{sec:ct_eval}. During this time, \mabbr{} made 221 true positive observations. Exemplary samples are visualized in figure \ref{fig:data_samples}. Each observation results in an HEVC encoded video with 30 frames per second and a length of 25 seconds. For our experiments, we sample two sets of still images from these videos.

For our camera trap distance sampling study (c.f. section \ref{sec:ctds}), we sample still images from the videos at a rate of $\SI{2}{\per\second}$, resulting in a total of $2871$ images.

For the instance segmentation task (c.f. section \ref{sec:insseg}), we want our dataset to consist of diverse scene configurations (animal positions and poses, lighting conditions, etc.). To obtain such diverse samples, sampling at regular intervals is not enough. Sometimes deer will stand still for long periods of time while moving quickly through the scene at other times. Therefore, we employ an approach based on background modeling using Gaussian mixture models \citep{mog_background_subtraction}. We then accumulate the ratio of foreground pixels (that is, the ratio of pixels occupied by moving objects) in each video frame until a threshold of 10\% is reached. This way, we sample more often if there is more movement in the video, and less often for less movement. We then annotated a total of 546 instances in 187 of the still images sampled this way with instance masks using the interactive annotation tool proposed by \cite{ritm}. Figure \ref{fig:labeling_samples} visualizes one such annotated image. On average, we needed roughly $3.5$ minutes per instance, resulting in a total annotation effort of roughly $32$ hours. Out of the total 546, we use 395 instances for training and validation (via 10-fold cross-validation), and reserve 151 instances as test dataset, such that images from a single video are only ever contained in one dataset. The test dataset is not used in this work but is instead reserved for future work. We publish both the raw data \citep{plittersdorf_dataset_raw} and the instance segmentation dataset \citep{plittersdorf_dataset_insseg}.

\subsection{Depth-aware Instance Segmentation}\label{sec:insseg}

We frame the problem of detecting and localizing animals as an instance segmentation problem, with the goal of generating a bounding box and a binary mask for each animal instance. Compared to animal presence-absence classification, this approach allows both counting the exact number of animals present, as well as inferring the distance between animal and camera by applying the binary mask to the depth images obtained using stereo vision. However, the depth images themselves obtain useful information for differentiating multiple individual animals from themselves and the background. Instance segmentation is usually performed by first forwarding a color image with red, green and blue (RGB) channels through the backbone, which may be a convolutional neural network (CNN) or a vision transformer \citep{vit}. The backbone then generates a hierarchy of feature maps that encode higher- and higher-level information about the image. These feature maps are then used by an instance segmentation model to predict bounding boxes and binary masks themselves. It is not obvious how to use the depth images obtained from \mabbr{} in this framework. Compared to datasets like ShapeNet \citep{shapenet}, we only have information from (effectively) a single perspective. We therefore argue that it is wise to treat the depth information as an additional channel in the two-dimensional image instead of working on point clouds or voxel grids, which increase the computational and memory requirements while largely foregoing the significant improvements being made in the area of 2D instance segmentation.

\begin{figure}[H]
\centering
\includegraphics[width=\textwidth]{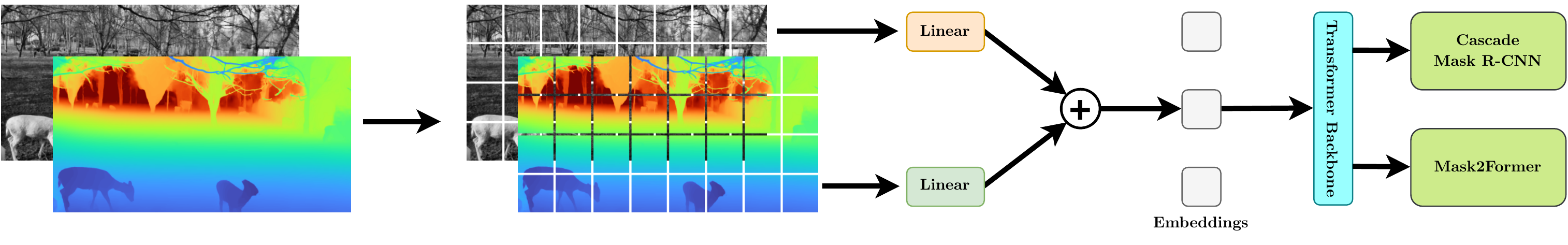}
\caption{RGB-D Instance Segmentation using Omnivore \citep{omnivore} and Cascade Mask R-CNN \citep{cmrcnn} or Mask2Former \citep{mask2former}. In Omnivore, the grayscale and depth images are first split into 2D patches, linearly embedded and added together. The resulting embeddings are then passed through a transformer encoder which generates hierarchical feature maps. These feature maps are then used by Cascade Mask R-CNN or Mask2Former to perform instance segmentation.}\label{fig:insseg_architecture}
\end{figure}

Although most backbones are pre-trained on color images without depth information, a recent work proposes \emph{Omnivore}, a vision transformer backbone trained on color and depth information \citep{omnivore}. In our experiments, we use \emph{Omnivore} as our backbone of choice. As instance segmentation models, we use either the convolution-based \emph{Cascade Mask R-CNN} \citep{cmrcnn} or the transformer-based \emph{Mask2Former} \citep{mask2former}. This is motivated by the observation that vision transformers require more training data to perform well, compared to CNNs \citep{vit,vit_big_data_1}. Therefore, Mask2Former performs very poorly when trained on small datasets such as the \emph{Plittersdorf instance segmentation dataset}, while outperforming Cascade Mask R-CNN on larger datasets such as Cityscapes \citep{cityscapes}. The resulting model architecture is visualized by figure \ref{fig:insseg_architecture}. To demonstrate that depth information is not only beneficial on the Plittersdorf instance segmentation task, we also evaluate improvements on the Cityscapes instance segmentation dataset. This is because the Cityscapes dataset is one of the only datasets which provides both depth information through stereo vision and a large amount of instance segmentation annotations.

We implement our instance segmentation pipeline using \emph{mmdetection} \citep{mmdetection} and largely keep the default hyperparameters of the mmdetection model implementations. We use the AdamW optimzer \citep{adamw} with a global learning rate of $5 N 10^{-5}$ with batch size $N$ and a weight decay of $0.05$. We set $N=2$ for Mask2Former and $N=6$ for Cascade Mask R-CNN due to memory constraints.

\subsection{Camera Trap Distance Sampling Study}\label{sec:ctds}
It is now possible to combine the instance masks generated by the animal detection model (c.f. section \ref{sec:insseg}) with the depth images obtained by stereo vision (c.f. section \ref{sec:stereo}) to obtain the distances required for the camera trap distance sampling (CTDS, \cite{camera_trap_distance_sampling}) abundance estimation method. To be able to use all observations without leaking information from the training dataset of our instance segmentation model, in this study, we use not the instance masks but the bounding boxes of MegaDetector \citep{megadetector} and the sampling approach by \cite{haucke2022overcoming}. To show the viability of this approach, we perform an exemplary estimation of detection probability.
We use 7 equally spaced distance intervals from $\SI{3}{\meter}$ to $\SI{11}{\meter}$. As \mabbr{} is mounted on a tree just outside the enclosure and at a height of $\SI{1.9}{\meter}$, $\SI{3}{\meter}$ is the minimal distance where deer are certain to be visible. We do not re-scale the minimum distance, as deer might be present closer to the camera but outside the field of view.
We use the \emph{Distance for Windows} software (version 7.4, \cite{distance_software}) and model the detection function using a uniform key function with a single cosine adjustment term.

\section{Evaluation}
\subsection[SOCRATES]{\mabbr{}}\label{sec:ct_eval}

We operated \mabbr{} in the \plittersdorf{}, Bonn, Germany, from February 9th to July 8th 2022, or 149 days. The \plittersdorf{} houses exclusively European fallow deer (Dama dama) and Sika deer (Cervus nippon). The camera was mounted on the side of a tree using a lashing strap and not moved during the entire duration. During this time, \mabbr{} experienced temperatures from $\SI{-4}{\celsius}$ to $\SI{38}{\celsius}$ and storms with wind speeds of $\SI{87}{\kilo\meter\per\hour}$ without issues. \mabbr{} was without power or the software disabled due to maintenance for 46 days, resulting in a total number of 103 observation days. During this time, \mabbr{} recorded 1089 observations. Out of these, 221 showed visible animals. This indicates a false positive rate of roughly $80\%$, which is in line with prior work concerned with commercial camera traps \citep{camera_trap_limitations}. False triggers are primarily induced by (1) animals in the field-of-view of the PIR sensor, but outside of the fields of view of the cameras, (2) excessive infrared illumination by the sun during daytime or (3) by artificial light sources such as flood lights on nearby buildings. We manually removed all false positive observations from our dataset. Although this could easily be automated, e.g. by using the MegaDetector \citep{megadetector}, we wanted to ensure that there are no persons in the final dataset and therefore screened the entire dataset manually.

\begin{table}[]
\centering
\begin{tabular}{@{}lcc@{}}
\toprule
                 & HP2XC \citep{reconyx_hp2xc}      & \mabbr{}               \\ \midrule
Provides Depth   & \xmark                 & \cmark                \\
Image Resolution & $1920 \times 1080$ / $3$MP  & $1920 \times 1080$        \\
Video Resolution & $1280\times 720$   & $1920 \times 1080$ \\
Video Length     & max. $\SI{90}{\second}$ at 2FPS      & up to $\SI{40}{\hour}$ at 30FPS \\
Daytime Imaging     & RGB         & near infrared                 \\
Nighttime Imaging     & near infrared         & near infrared                 \\
Illumination Wavelength     & $\SI{940}{\nano\meter}$         & $\SI{850}{\nano\meter}$                 \\
Dimensions       & $14 \times 11.5 \times \SI{7.5}{\centi\meter}$ & $11.6 \times 80 \times \SI{20}{\centi\meter}$    \\
Connectivity     & Cellular           & Cellular / W-LAN / LAN   \\
Battery Life     & up to a year             & $\sim$ 9 days        \\
Material Cost             & \$659.99           & $\sim$ \$900       \\ \bottomrule
\end{tabular}
\caption{Comparison between \mabbr{} and the widely used Reconyx HP2XC trail camera. The video length of \mabbr{} is only bounded by the available persistent storage space, although we use $\SI{25}{\second}$ videos for our experiments. Costs include only materials and not assembly.}
\label{tab:commercial_comparison}
\end{table}

We compare \mabbr{} with the widely used commercially produced Reconyx HP2XC in Table \ref{tab:commercial_comparison}. \mabbr{} is significantly larger to support large baselines, while having significantly shorter battery life and slightly higher component costs. The infrared illuminator of \mabbr{} operates at a slightly shorter wavelength, which might be visible for some animals. The infrared illuminator should therefore be replaced with a longer-wavelength version in the future. At the same time, \mabbr{} not only provides depth information through stereo vision, but also allows recording video at high resolutions and frame rates for long durations only limited by available storage space.

We demonstrate that the stereo capabilities \mabbr{} facilitate improved visual animal detection (c.f. section \ref{sec:animal_detection_eval}) and accurate abundance estimation using camera trap distance sampling (c.f. section \ref{sec:ctds_eval}). Depth information is also essential for obtaining absolute animal sizes in photogrammetry, which is traditionally performed using laser rangefinders \citep{shrader2006digital}. Furthermore, depth information has been shown to improve the accuracy of animal tracking over 2D-only approaches \citep{morris_tracking_1,morris_tracking_2}. \mabbr{} can not compete with commercially available camera traps in cost or battery life, but this is not our goal. Apart from the methodological improvements described above, \mabbr{} fulfills three high-level goals:
\begin{enumerate}
    \item it demonstrates that stereo camera traps are viable and worthwhile. We hope to convince commercial camera trap manufacturers to support stereo camera setups using off-the-shelf hardware.
    \item it facilitates the verification of monocular approaches. For example, abundance estimation using camera trap distance sampling might be performed twice, once using monocular approaches \citep{haucke2022overcoming,johanns2022automated} and once using \mabbr{}. Both raw animal distances and the resulting animal densities might then be compared.
    \item it allows to generate training data for monocular depth estimation methods such as \cite{monodepth2,midas,dpt}. These approaches have been largely focused on human-centric scenes such as indoor and street scenes with relatively simple geometry, which are highly unlike natural scenes such as forests. Gathering training data from natural scenes might help these methods to generalize better to such scenes and thus allow monocular camera traps to more accurately estimate depth information in the future.
\end{enumerate}

\subsubsection{Stereo Correspondence}

Figure \ref{fig:data_samples} shows some exemplary pairs of near infrared images and corresponding depth maps inferred by \citep{crestereo}. As can be seen, the depth maps generally represent the scene well and clearly highlight the boundaries of the deer. To evaluate the depth maps quantitatively, we employ the temporal quality metric proposed in \citep{disparity_map_quality}, which is defined as:
\begin{equation}
    E_t = \frac{1}{(N_T - 1)N_P} \sum_{n=2}^{N_T} \sum_{(x, y)} \lvert D(x, y, n) - D(x - m_x, y - m_y, n - 1) \rvert
\end{equation}
where $N_T$ is equal to the number of frames in the input video, $N_P$ is the number of pixels in a single frame, $D(x, y, n)$ is the scalar disparity at some pixel $(x, y)$ at time $n$, and $m_x, m_y$ is the optical flow from frame $n$ to frame $n - 1$, calculated using \citep{farneback_optical_flow}. Using \citep{crestereo}, we obtain $E_t = 0.4439$, which is on-par with the temporal error of the ground truth disparity in \citep{disparity_map_quality}. 

\begin{wrapfigure}{r}{0.5\textwidth}
  \vspace{-0.25cm}
  \begin{center}
    \includegraphics[width=0.48\textwidth]{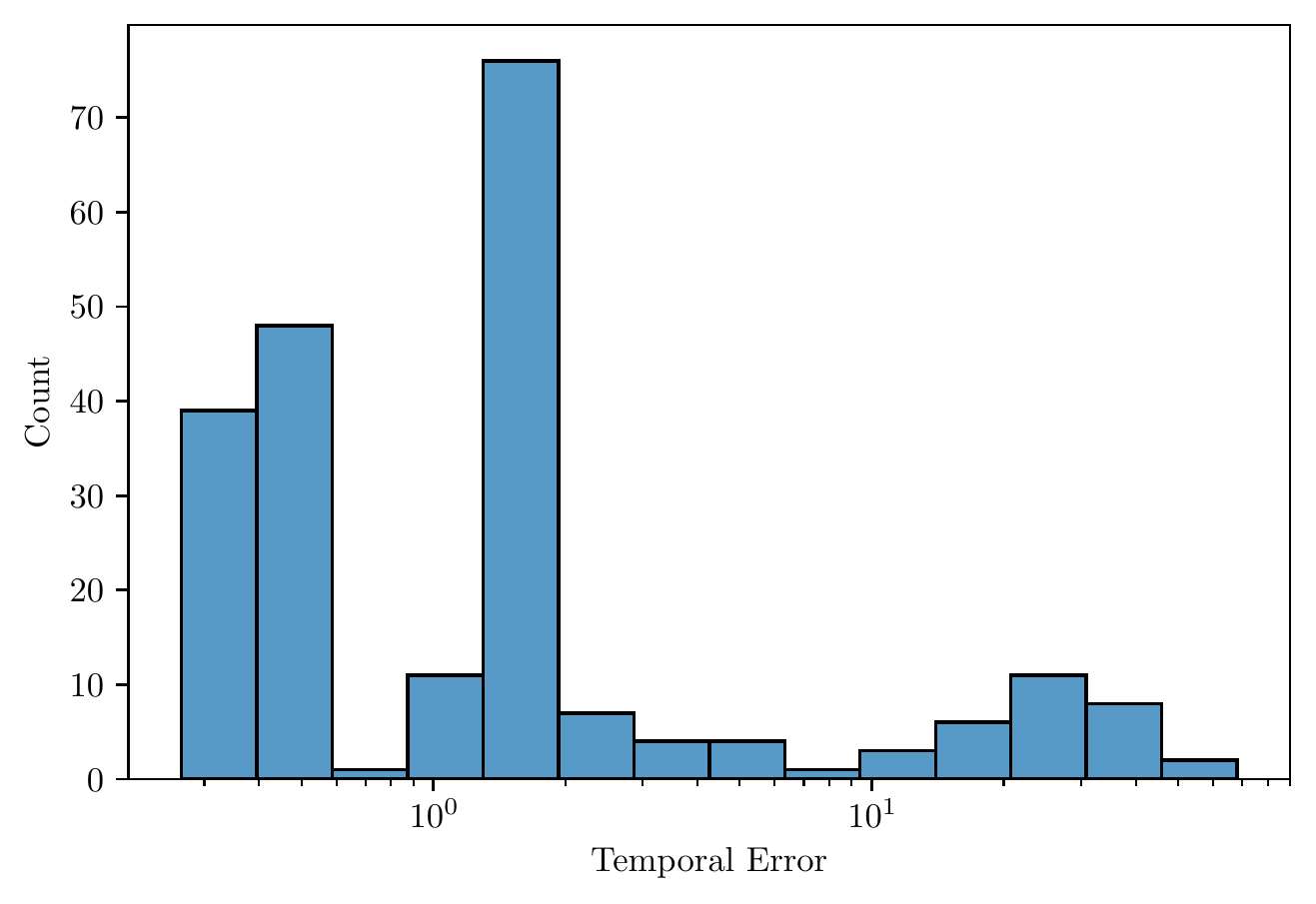}
  \end{center}
  \caption{Histogram plot of the mean per-observation temporal stereo correspondence error.}
  \label{fig:temporal_error_histogram}
\end{wrapfigure}

\begin{figure}
     \centering
     \begin{subfigure}[b]{\textwidth}
         \centering
         \includegraphics[width=\textwidth]{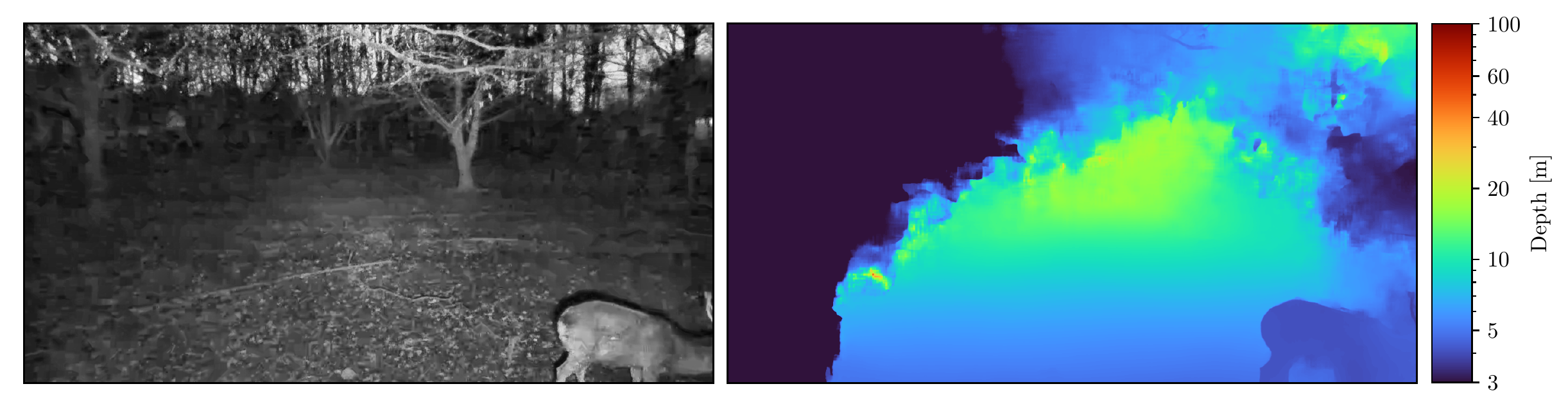}
     \end{subfigure}
    \caption{Stereo matching inevitably fails in regions where there is not enough available information.}
    \label{fig:stereo_failure_samples}
\end{figure}

As can be seen, the temporal error is low for the vast majority of observations. Like with regular camera traps, at night time, some regions in the field of view might be insufficiently lit and therefore underexposed in the resulting images. In these regions, insufficient image information is available to perform successful stereo correspondence, leading to the outliers with poor temporal error apparent in figure \ref{fig:temporal_error_histogram}. One such outlier case is shown in figure \ref{fig:stereo_failure_samples}. Still, the depth of the well-lit area is correctly inferred.

\subsection{Visual Animal Detection}\label{sec:animal_detection_eval}
\begin{table}[]
\resizebox{\textwidth}{!}{
\begin{tabular}{@{}llllllllll@{}}
\toprule
Backbone                                & $\text{AP}^\text{bbox}$ & $\text{AP}^\text{bbox}_{50}$ & $\text{AP}^\text{bbox}_{75}$ & $\text{AP}^\text{segm}$ & $\text{AP}^\text{segm}_{50}$ & $\text{AP}^\text{segm}_{75}$ & $\text{AP}^\text{segm}_\text{l}$ & $\text{AP}^\text{segm}_\text{m}$ & $\text{AP}^\text{segm}_\text{s}$ \\ \midrule
Swin-L                            & 0.5164                  & 0.8438                       & 0.5272                       & 0.4359                  & 0.8328                       & 0.4285                       & 0.5192                           & 0.3856                           & 0.1285                           \\
Omnivore-L   & 0.5243                  & \textbf{0.8598}              & 0.5702                       & 0.4382                  & \textbf{0.8353}              & 0.4376                       & 0.5150                           & 0.3895                           & \textbf{0.1431}                  \\
Depth-aware Omnivore-L & \textbf{0.5399}         & 0.8494                       & \textbf{0.6048}              & \textbf{0.4547}         & 0.8147                       & \textbf{0.4699}              & \textbf{0.5427}                  & \textbf{0.4013}                  & 0.1138                           \\ \bottomrule
\end{tabular}
}
\caption{COCO metrics on the Plittersdorf Instance Segmentation Dataset task using Cascade Mask R-CNN with different backbones and 10-fold cross-validation.}
\label{tab:plittersdorf_eval}
\end{table}
We use the COCO \citep{coco} metrics to evaluate our instance segmentation models. Each metric is obtained by performing 10-fold cross-validation after the last training epoch. Cross-validation is especially important in this setting, as it reduces the impact of a single lucky train-test split on this small dataset. Table \ref{tab:plittersdorf_eval} summarizes the results on the Plittersdorf instance segmentation task. The summarizing metrics for bounding boxes ($\text{AP}^\text{bbox}$) and segmentation ($\text{AP}^\text{segm}$) show that incorporating depth information results in an overall performance improvement. Interestingly, for low IOU thresholds ($\text{AP}^\text{bbox}_{50}$, $\text{AP}^\text{segm}_{50}$), depth information seems to have the opposite effect. In other words, pure grayscale images perform better for roughly localizing an animal, whereas grayscale and depth information together are better for localizing animals very accurately $\text{AP}^\text{bbox}_{75}$, $\text{AP}^\text{segm}_{75}$). This is especially interesting as the ground truth labeling is performed using exclusively the grayscale image. Intuitively, one could therefore argue that the grayscale information is most important for matching the ground truth very precisely. Here we see the opposite effect. As the error of stereo correspondence is quadratically related to the true distance, the resulting depth maps become less useful at larger distances. This is reflected in the lower performance on small instances ($\text{AP}^\text{segm}_{s}$), which typically are farther away than medium ($\text{AP}^\text{segm}_{m}$) or large instances ($\text{AP}^\text{segm}_{l}$). We tried to ease the dependence on depth information for these faraway instances by clipping the depth values to different maximum distances or randomly dropping the depth information altogether during training \citep{dropout}. However, this did not result in meaningful improvements.
\subsection{Depth-aware Instance Segmentation on Cityscapes}
\begin{table}[]
\resizebox{\textwidth}{!}{
\begin{tabular}{@{}lllllll@{}}
\toprule
Backbone               & $\text{AP}^\text{bbox}$ & $\text{AP}^\text{segm}$ & $\text{AP}^\text{segm}_{50}$ & $\text{AP}^\text{segm}_\text{l}$ & $\text{AP}^\text{segm}_\text{m}$ & $\text{AP}^\text{segm}_\text{s}$ \\ \midrule
Swin-L               &                         & 0.437                   & 0.714                        &                                  & \textbf{}                        &                                  \\
Omnivore-L             & 0.415                   & 0.439                   & 0.700                        & 0.716                            & 0.394                            & 0.214                            \\
Depth-aware Omnivore-L & \textbf{0.431}          & \textbf{0.456}          & \textbf{0.734}               & \textbf{0.732}                   & \textbf{0.411}                   & \textbf{0.264}                   \\ \bottomrule
\end{tabular}
}
\caption{Instance segmentation results on the Cityscapes validation set. The depth-aware Omnivore-L variant clearly improves the non-depth-aware variant in all metrics. The metrics of the Swin-L backbone are obtained using the original implementation \citep{mask2former}. $\text{AP}^\text{segm}$ and $\text{AP}^\text{segm}_{50}$ are Cityscapes metrics, the rest are COCO metrics.}\label{tab:cityscapes_eval}
\end{table}
To show that the positive effect of depth information on instance segmentation accuracy is not limited to settings with grayscale images, a single object class, and a fixed camera such as \mabbr{}, we additionally evaluate our instance segmentation approach on the Cityscapes instance segmentation task \citep{cityscapes}. The Cityscapes instance segmentation dataset \citep{cityscapes} is composed of color and stereo depth images of urban street scenes, captured by cameras in a moving car. It features several object classes, such as \emph{person}, \emph{car}, or \emph{bus}, annotated with instance labels. The Cityscapes dataset is also much larger, with 3475 annotated images in its train and validation sets. As can be seen in table \ref{tab:cityscapes_eval}, the depth information has an overall even greater positive impact than in the Plittersdorf task (c.f. section \ref{sec:animal_detection_eval}). This is likely caused by two reasons: (1) the Mask2Former \citep{mask2former} being able to better make use of the depth-aware feature hierarchies produced by Omnivore \citep{omnivore}, and (2), the larger training dataset, which might help alleviate the lower number of depth images during pre-training \citep{omnivore}.
%
%
\subsection[Abundance Estimation using SOCRATES]{Abundance Estimation using \mabbr{}}\label{sec:ctds_eval}

%
\begin{wrapfigure}{l}{0.48\textwidth}
  \vspace{-0.5cm}
  \begin{center}
    \includegraphics[width=0.48\textwidth]{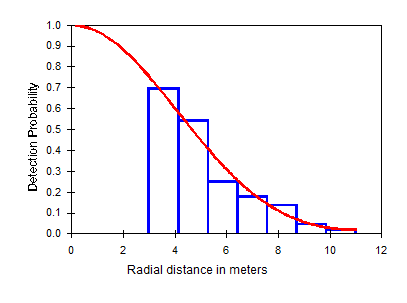}
  \end{center}
  \vspace{-0.5cm}
  \caption{CTDS detection probability. We transform our distance measurements into seven intervals (visualized in blue) from which the detection probability (visualized in red) is derived using CTDS.}
  \label{fig:ctds_pdf}
\end{wrapfigure}
Figure \ref{fig:ctds_pdf} depicts the detection probability obtained by CTDS using the parameters specified in section \ref{sec:ctds}.
%
%
%
Note that the estimated probability density approximates the measurements well, starting from a distance of $\SI{3}{\meter}$. Due to the way \mabbr{} is mounted, deer below $\SI{3}{\meter}$ may not be visible, which is why we exclude these low distances from our estimation (c.f. section \ref{sec:ctds}).
\begin{figure}[h!]
    \centering
    \includegraphics[width=0.75\textwidth]{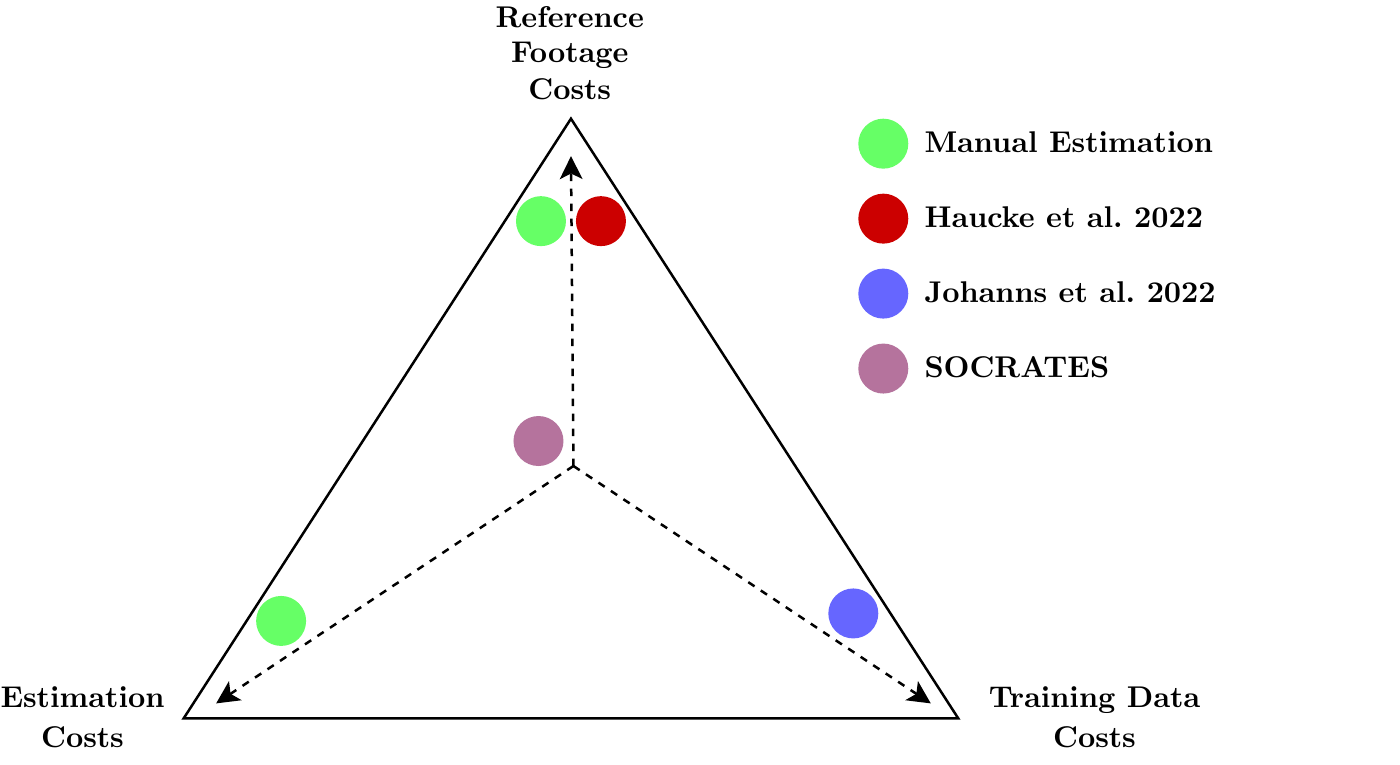}
    \caption{Tradeoff between reference footage costs, manual distance estimation costs and training data costs. The manual approach requires both the acquisition of reference footage and labour-intensive manual animal distance estimation. \cite{haucke2022overcoming} require reference footage, but the distance estimation itself is automated. \cite{johanns2022automated} require neither reference footage nor manual distance estimations, but requires training data of similar scenes. \mabbr{} measures distance using stereo vision and therefore incurs none of these costs.}
    \label{fig:tradeoff}
\end{figure}
Compared to competing approaches \citep{haucke2022overcoming,johanns2022automated}, distance estimation for abundance estimation of unmarked animal populations is straightforward with \mabbr{}. Figure \ref{fig:tradeoff} visualizes the respective tradeoffs.
The presented proof-of-concept for modelling detection probability in camera trap distance sampling with \mabbr{} stereo camera devices demonstrates the usefulness of the approaches taken and its potential for improving the efficacy of future wildlife surveys. The reduced cost for data processing, the increase in animal detection and potential for application in integrated mono- and stereo-camera trap surveys pave the way for an end-to-end solution in computational wildlife monitoring. The proposed approach is not limited to the conditions of our study, but is widely applicable across habitats, species and regions. For future field surveys, we recommend that multiple \mabbr{} devices be used, along with a random or systematic study design to estimate wildlife density and associated variance reliably. \mabbr{} can  also be paired cooperatively with traditional, monocular camera traps for improved error quantification and improvement of monocular distance estimations like that proposed by \cite{johanns2022automated}.
%
%
%
\subsection{AMMOD Portal Case Study}\label{sec:ammod_portal}
\begin{figure}
     \centering
     \begin{subfigure}[b]{\textwidth}
         \centering
         \includegraphics[width=\textwidth]{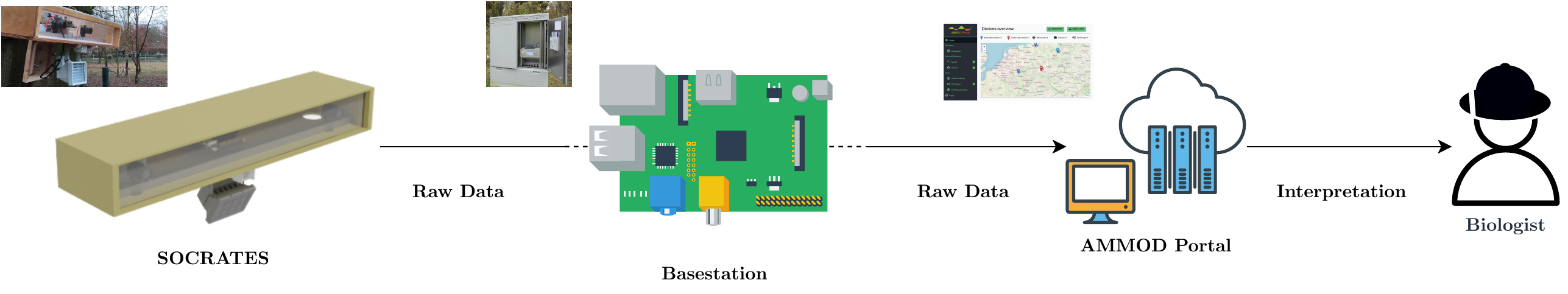}
     \end{subfigure}
    \caption{Fully automatic flow of data from \mabbr{} over the basestation to the AMMOD Portal and the expert end users. A GPU server runs the instance segmentation and distance estimation steps, and uploads the results back to the AMMOD portal.}
    \label{fig:ammod_workflow}
\end{figure}

A central goal of the AMMOD project is to automatically collect all observed data in a central repository (the \emph{AMMOD Portal}, \href{https://data.ammod.de}{https://data.ammod.de}), which will eventually be accessible to biologists and the general public. For \mabbr{}, we ensure this by uploading the captured raw data via the CoAP protocol \citep{coap} to the \emph{AMMOD Basestation} \citep{waegele_ammod,basestation}, if available at the current location, or directly to the AMMOD Portal otherwise. The AMMOD Basestation takes the role of scheduling and prioritizing data transfer from different sensors according to the energy available from energy harvesting. Once the raw data is uploaded to the AMMOD Portal, a server runs the instance segmentation (c.f. section \ref{sec:insseg}) and distance estimation (c.f. section \ref{sec:ctds}) workflows. To increase throughput and energy efficiency, the server is equipped with an NVIDIA GPU to accelerate neural network inference. Both methods are packaged as \emph{Docker} images to simplify dependency management and updates. The resulting instance masks and distances are then again uploaded to the AMMOD Portal and are available for further analysis by biologists. The entire data flow is fully automated and visualized in figure \ref{fig:ammod_workflow}.

\section{Conclusion}
%
\noindent
We propose a \mname{} (\mabbr{}), a novel camera trap prototype that uses stereo vision to infer the 3D structure of the captured scene. \mabbr{} enables the following contributions:

\noindent $\bullet$
\textit{\textbf{Detection and localization of animals}}: \mabbr{} provides depth information that improves the accurate localization of animals in an instance segmentation setting, e.g. by $3.23\%$ in bounding box $\text{mAP}_{75}$. We obtain these results by performing 10-fold cross-validation. Similar results on the Cityscapes instance segmentation task show that this effect is neither limited to grayscale images, a single object category, nor fixed cameras such as \mabbr{}.

\noindent $\bullet$
\textit{\textbf{Abundance estimation}} is facilitated by the automatic distance measurements of \mabbr{}. We perform a proof-of-concept camera trap distance sampling study and successfully model detection probability in a wildlife enclosure. Future work might use \mabbr{} to perform automatic abundance estimation in the wild and compare the results with competing approaches.

\noindent $\bullet$
\textit{\textbf{Reproducibility and accessibility}} for practitioners is enabled by openly providing our raw and labeled data, code, detailed instructions, best practices, and 3D CAD models at \href{https://github.com/timmh/socrates}{https://github.com/timmh/socrates}. We hope to pave the way for the eventual adaption of stereo camera traps by commercial manufacturers.




\clearpage

\noindent 
\section*{\centering \begin{normalsize}Acknowledgement\end{normalsize}}
This work is partially funded by the German Federal Ministry of Education and Research (Bundesministerium für Bildung und Forschung (BMBF), Bonn, Gemany (AMMOD - Automated Multisensor Stations for Monitoring of BioDiversity: FKZ 01LC1903B). This funding is gratefully acknowledged.

We thank Vincent Mainzer and the team of the \plittersdorf{} for their cooperation by hosting the camera trap hardware on-site.

\pagebreak

\Urlmuskip=0mu plus 1mu\relax
\bibliographystyle{elsarticle-harv}
\bibliography{references}

\begin{thebibliography}{51}
\expandafter\ifx\csname natexlab\endcsname\relax\def\natexlab#1{#1}\fi
\providecommand{\url}[1]{\texttt{#1}}
\providecommand{\href}[2]{#2}
\providecommand{\path}[1]{#1}
\providecommand{\DOIprefix}{doi:}
\providecommand{\ArXivprefix}{arXiv:}
\providecommand{\URLprefix}{URL: }
\providecommand{\Pubmedprefix}{pmid:}
\providecommand{\doi}[1]{\href{http://dx.doi.org/#1}{\path{#1}}}
\providecommand{\Pubmed}[1]{\href{pmid:#1}{\path{#1}}}
\providecommand{\bibinfo}[2]{#2}
\ifx\xfnm\relax \def\xfnm[#1]{\unskip,\space#1}\fi
\bibitem[{{Altuglas International}(2000)}]{plexiglass_transmission}
\bibinfo{author}{{Altuglas International}}, \bibinfo{year}{2000}.
\newblock \bibinfo{title}{Plexiglas -- optical \& transmission
  characteristics}.
\newblock \bibinfo{journal}{Arkema, Philadelphia, PA} .
\bibitem[{Auda(2022)}]{auda_johanns_issues}
\bibinfo{author}{Auda, E.}, \bibinfo{year}{2022}.
\newblock \bibinfo{title}{Overestimation of animal distances in close-up
  scenarios}.
\newblock \bibinfo{howpublished}{Wildlife Conservation Society - Cambodia}.
\newblock \bibinfo{note}{Personal communication}.
\bibitem[{Beery et~al.(2019)Beery, Morris and Yang}]{megadetector}
\bibinfo{author}{Beery, S.}, \bibinfo{author}{Morris, D.},
  \bibinfo{author}{Yang, S.}, \bibinfo{year}{2019}.
\newblock \bibinfo{title}{Efficient pipeline for camera trap image review}.
\newblock \bibinfo{journal}{arXiv preprint arXiv:1907.06772} .
\bibitem[{Bormann et~al.(2012)Bormann, Castellani and Shelby}]{coap}
\bibinfo{author}{Bormann, C.}, \bibinfo{author}{Castellani, A.P.},
  \bibinfo{author}{Shelby, Z.}, \bibinfo{year}{2012}.
\newblock \bibinfo{title}{Coap: An application protocol for billions of tiny
  internet nodes}.
\newblock \bibinfo{journal}{IEEE Internet Computing} \bibinfo{volume}{16},
  \bibinfo{pages}{62--67}.
\newblock \DOIprefix\doi{10.1109/MIC.2012.29}.
\bibitem[{Cai and Vasconcelos(2018)}]{cmrcnn}
\bibinfo{author}{Cai, Z.}, \bibinfo{author}{Vasconcelos, N.},
  \bibinfo{year}{2018}.
\newblock \bibinfo{title}{Cascade r-cnn: Delving into high quality object
  detection}, in: \bibinfo{booktitle}{Proceedings of the IEEE conference on
  computer vision and pattern recognition}, pp. \bibinfo{pages}{6154--6162}.
\bibitem[{Chang et~al.(2015)Chang, Funkhouser, Guibas, Hanrahan, Huang, Li,
  Savarese, Savva, Song, Su, Xiao, Yi and Yu}]{shapenet}
\bibinfo{author}{Chang, A.X.}, \bibinfo{author}{Funkhouser, T.},
  \bibinfo{author}{Guibas, L.}, \bibinfo{author}{Hanrahan, P.},
  \bibinfo{author}{Huang, Q.}, \bibinfo{author}{Li, Z.},
  \bibinfo{author}{Savarese, S.}, \bibinfo{author}{Savva, M.},
  \bibinfo{author}{Song, S.}, \bibinfo{author}{Su, H.}, \bibinfo{author}{Xiao,
  J.}, \bibinfo{author}{Yi, L.}, \bibinfo{author}{Yu, F.},
  \bibinfo{year}{2015}.
\newblock \bibinfo{title}{{ShapeNet: An Information-Rich 3D Model Repository}}.
\newblock \bibinfo{type}{Technical Report} \bibinfo{number}{arXiv:1512.03012
  [cs.GR]}. Stanford University --- Princeton University --- Toyota
  Technological Institute at Chicago.
\bibitem[{Chen et~al.(2019)Chen, Wang, Pang, Cao, Xiong, Li, Sun, Feng, Liu,
  Xu, Zhang, Cheng, Zhu, Cheng, Zhao, Li, Lu, Zhu, Wu, Dai, Wang, Shi, Ouyang,
  Loy and Lin}]{mmdetection}
\bibinfo{author}{Chen, K.}, \bibinfo{author}{Wang, J.}, \bibinfo{author}{Pang,
  J.}, \bibinfo{author}{Cao, Y.}, \bibinfo{author}{Xiong, Y.},
  \bibinfo{author}{Li, X.}, \bibinfo{author}{Sun, S.}, \bibinfo{author}{Feng,
  W.}, \bibinfo{author}{Liu, Z.}, \bibinfo{author}{Xu, J.},
  \bibinfo{author}{Zhang, Z.}, \bibinfo{author}{Cheng, D.},
  \bibinfo{author}{Zhu, C.}, \bibinfo{author}{Cheng, T.},
  \bibinfo{author}{Zhao, Q.}, \bibinfo{author}{Li, B.}, \bibinfo{author}{Lu,
  X.}, \bibinfo{author}{Zhu, R.}, \bibinfo{author}{Wu, Y.},
  \bibinfo{author}{Dai, J.}, \bibinfo{author}{Wang, J.}, \bibinfo{author}{Shi,
  J.}, \bibinfo{author}{Ouyang, W.}, \bibinfo{author}{Loy, C.C.},
  \bibinfo{author}{Lin, D.}, \bibinfo{year}{2019}.
\newblock \bibinfo{title}{{MMDetection}: Open mmlab detection toolbox and
  benchmark}.
\newblock \bibinfo{journal}{arXiv preprint arXiv:1906.07155} .
\bibitem[{Cheng et~al.(2022)Cheng, Misra, Schwing, Kirillov and
  Girdhar}]{mask2former}
\bibinfo{author}{Cheng, B.}, \bibinfo{author}{Misra, I.},
  \bibinfo{author}{Schwing, A.G.}, \bibinfo{author}{Kirillov, A.},
  \bibinfo{author}{Girdhar, R.}, \bibinfo{year}{2022}.
\newblock \bibinfo{title}{Masked-attention mask transformer for universal image
  segmentation}.
\bibitem[{Cordts et~al.(2016)Cordts, Omran, Ramos, Rehfeld, Enzweiler,
  Benenson, Franke, Roth and Schiele}]{cityscapes}
\bibinfo{author}{Cordts, M.}, \bibinfo{author}{Omran, M.},
  \bibinfo{author}{Ramos, S.}, \bibinfo{author}{Rehfeld, T.},
  \bibinfo{author}{Enzweiler, M.}, \bibinfo{author}{Benenson, R.},
  \bibinfo{author}{Franke, U.}, \bibinfo{author}{Roth, S.},
  \bibinfo{author}{Schiele, B.}, \bibinfo{year}{2016}.
\newblock \bibinfo{title}{The cityscapes dataset for semantic urban scene
  understanding}, in: \bibinfo{booktitle}{Proceedings of the IEEE Conference on
  Computer Vision and Pattern Recognition (CVPR)}.
\bibitem[{Deng et~al.(2009)Deng, Dong, Socher, Li, Li and Fei-Fei}]{imagenet}
\bibinfo{author}{Deng, J.}, \bibinfo{author}{Dong, W.},
  \bibinfo{author}{Socher, R.}, \bibinfo{author}{Li, L.J.},
  \bibinfo{author}{Li, K.}, \bibinfo{author}{Fei-Fei, L.},
  \bibinfo{year}{2009}.
\newblock \bibinfo{title}{{ImageNet: A Large-Scale Hierarchical Image
  Database}}, in: \bibinfo{booktitle}{CVPR09}.
\bibitem[{Dosovitskiy et~al.(2020)Dosovitskiy, Beyer, Kolesnikov, Weissenborn,
  Zhai, Unterthiner, Dehghani, Minderer, Heigold, Gelly et~al.}]{vit}
\bibinfo{author}{Dosovitskiy, A.}, \bibinfo{author}{Beyer, L.},
  \bibinfo{author}{Kolesnikov, A.}, \bibinfo{author}{Weissenborn, D.},
  \bibinfo{author}{Zhai, X.}, \bibinfo{author}{Unterthiner, T.},
  \bibinfo{author}{Dehghani, M.}, \bibinfo{author}{Minderer, M.},
  \bibinfo{author}{Heigold, G.}, \bibinfo{author}{Gelly, S.}, et~al.,
  \bibinfo{year}{2020}.
\newblock \bibinfo{title}{An image is worth 16x16 words: Transformers for image
  recognition at scale}.
\newblock \bibinfo{journal}{arXiv preprint arXiv:2010.11929} .
\bibitem[{Farneb{\"a}ck(2003)}]{farneback_optical_flow}
\bibinfo{author}{Farneb{\"a}ck, G.}, \bibinfo{year}{2003}.
\newblock \bibinfo{title}{Two-frame motion estimation based on polynomial
  expansion}, in: \bibinfo{booktitle}{Scandinavian conference on Image
  analysis}, \bibinfo{organization}{Springer}. pp. \bibinfo{pages}{363--370}.
\bibitem[{Girdhar et~al.(2022)Girdhar, Singh, Ravi, van~der Maaten, Joulin and
  Misra}]{omnivore}
\bibinfo{author}{Girdhar, R.}, \bibinfo{author}{Singh, M.},
  \bibinfo{author}{Ravi, N.}, \bibinfo{author}{van~der Maaten, L.},
  \bibinfo{author}{Joulin, A.}, \bibinfo{author}{Misra, I.},
  \bibinfo{year}{2022}.
\newblock \bibinfo{title}{{Omnivore: A Single Model for Many Visual
  Modalities}}, in: \bibinfo{booktitle}{CVPR}.
\bibitem[{Godard et~al.(2019)Godard, {Mac Aodha}, Firman and
  Brostow}]{monodepth2}
\bibinfo{author}{Godard, C.}, \bibinfo{author}{{Mac Aodha}, O.},
  \bibinfo{author}{Firman, M.}, \bibinfo{author}{Brostow, G.J.},
  \bibinfo{year}{2019}.
\newblock \bibinfo{title}{Digging into self-supervised monocular depth
  prediction} .
\bibitem[{Hassani et~al.(2021)Hassani, Walton, Shah, Abuduweili, Li and
  Shi}]{vit_big_data_1}
\bibinfo{author}{Hassani, A.}, \bibinfo{author}{Walton, S.},
  \bibinfo{author}{Shah, N.}, \bibinfo{author}{Abuduweili, A.},
  \bibinfo{author}{Li, J.}, \bibinfo{author}{Shi, H.}, \bibinfo{year}{2021}.
\newblock \bibinfo{title}{Escaping the big data paradigm with compact
  transformers}.
\newblock \bibinfo{journal}{CoRR} \bibinfo{volume}{abs/2104.05704}.
\newblock \URLprefix \url{https://arxiv.org/abs/2104.05704},
  \href{http://arxiv.org/abs/2104.05704}{{\tt arXiv:2104.05704}}.
\bibitem[{Haucke et~al.(2022)Haucke, Kühl, Hoyer and
  Steinhage}]{haucke2022overcoming}
\bibinfo{author}{Haucke, T.}, \bibinfo{author}{Kühl, H.S.},
  \bibinfo{author}{Hoyer, J.}, \bibinfo{author}{Steinhage, V.},
  \bibinfo{year}{2022}.
\newblock \bibinfo{title}{Overcoming the distance estimation bottleneck in
  estimating animal abundance with camera traps}.
\newblock \bibinfo{journal}{Ecological Informatics} \bibinfo{volume}{68},
  \bibinfo{pages}{101536}.
\newblock \URLprefix
  \url{https://www.sciencedirect.com/science/article/pii/S1574954121003277},
  \DOIprefix\doi{https://doi.org/10.1016/j.ecoinf.2021.101536}.
\bibitem[{Haucke and Steinhage(2022a)}]{plittersdorf_dataset_insseg}
\bibinfo{author}{Haucke, T.}, \bibinfo{author}{Steinhage, V.},
  \bibinfo{year}{2022}a.
\newblock \bibinfo{title}{{SOCRATES Plittersdorf Instance Segmentation
  Dataset}}.
\newblock \URLprefix \url{https://doi.org/10.5281/zenodo.7035934},
  \DOIprefix\doi{10.5281/zenodo.7035934}.
\bibitem[{Haucke and Steinhage(2022b)}]{plittersdorf_dataset_raw}
\bibinfo{author}{Haucke, T.}, \bibinfo{author}{Steinhage, V.},
  \bibinfo{year}{2022}b.
\newblock \bibinfo{title}{{SOCRATES Plittersdorf Raw Data}}.
\newblock \URLprefix \url{https://doi.org/10.5281/zenodo.6992653},
  \DOIprefix\doi{10.5281/zenodo.6992653}.
\bibitem[{Howe et~al.(2017)Howe, Buckland, Després-Einspenner and
  Kühl}]{camera_trap_distance_sampling}
\bibinfo{author}{Howe, E.J.}, \bibinfo{author}{Buckland, S.T.},
  \bibinfo{author}{Després-Einspenner, M.L.}, \bibinfo{author}{Kühl, H.S.},
  \bibinfo{year}{2017}.
\newblock \bibinfo{title}{Distance sampling with camera traps}.
\newblock \bibinfo{journal}{Methods in Ecology and Evolution}
  \bibinfo{volume}{8}, \bibinfo{pages}{1558--1565}.
\newblock \URLprefix
  \url{https://besjournals.onlinelibrary.wiley.com/doi/abs/10.1111/2041-210X.12790},
  \DOIprefix\doi{10.1111/2041-210X.12790},
  \href{http://arxiv.org/abs/https://besjournals.onlinelibrary.wiley.com/doi/pdf/10.1111\\/2041-210X.12790}{{\tt
  arXiv:https://besjournals.onlinelibrary.wiley.com/doi/pdf/10.1111\\/2041-210X.12790}}.
\bibitem[{{Inayat Rasool}()}]{3d_cam}
\bibinfo{author}{{Inayat Rasool}}, .
\newblock \bibinfo{title}{Raspberry pi hq camera with arducam cs mount lens}.
\newblock
  \bibinfo{howpublished}{\url{https://grabcad.com/library/raspberry-pi-hq-camera-with-arducam-cs-mount-lens-1}}.
\bibitem[{Johanns et~al.(2022)Johanns, Haucke and
  Steinhage}]{johanns2022automated}
\bibinfo{author}{Johanns, P.}, \bibinfo{author}{Haucke, T.},
  \bibinfo{author}{Steinhage, V.}, \bibinfo{year}{2022}.
\newblock \bibinfo{title}{Automated distance estimation for wildlife camera
  trapping}.
\newblock \bibinfo{journal}{Ecological Informatics} \bibinfo{volume}{70},
  \bibinfo{pages}{101734}.
\newblock \URLprefix
  \url{https://www.sciencedirect.com/science/article/pii/S1574954122001844},
  \DOIprefix\doi{https://doi.org/10.1016/j.ecoinf.2022.101734}.
\bibitem[{{Juan Andres Viera Medina}()}]{3d_ir}
\bibinfo{author}{{Juan Andres Viera Medina}}, .
\newblock \bibinfo{title}{Infrared illuminator}.
\newblock
  \bibinfo{howpublished}{\url{https://grabcad.com/library/infrared-illuminator-1}}.
\bibitem[{KaewTraKulPong and Bowden(2002)}]{mog_background_subtraction}
\bibinfo{author}{KaewTraKulPong, P.}, \bibinfo{author}{Bowden, R.},
  \bibinfo{year}{2002}.
\newblock \bibinfo{title}{An improved adaptive background mixture model for
  real-time tracking with shadow detection}, in:
  \bibinfo{booktitle}{Video-based surveillance systems}.
  \bibinfo{publisher}{Springer}, pp. \bibinfo{pages}{135--144}.
\bibitem[{Klasen and Steinhage(2022a)}]{morris_tracking_1}
\bibinfo{author}{Klasen, M.}, \bibinfo{author}{Steinhage, V.},
  \bibinfo{year}{2022}a.
\newblock \bibinfo{title}{Improving wildlife tracking using 3d information}.
\newblock \bibinfo{journal}{Ecological Informatics} \bibinfo{volume}{68},
  \bibinfo{pages}{101535}.
\newblock \URLprefix
  \url{https://www.sciencedirect.com/science/article/pii/S1574954121003265},
  \DOIprefix\doi{https://doi.org/10.1016/j.ecoinf.2021.101535}.
\bibitem[{Klasen and Steinhage(2022b)}]{morris_tracking_2}
\bibinfo{author}{Klasen, M.}, \bibinfo{author}{Steinhage, V.},
  \bibinfo{year}{2022}b.
\newblock \bibinfo{title}{Wildlife 3d multi-object tracking}.
\newblock \bibinfo{journal}{Ecological Informatics} \bibinfo{volume}{71},
  \bibinfo{pages}{101790}.
\newblock \URLprefix
  \url{https://www.sciencedirect.com/science/article/pii/S1574954122002400},
  \DOIprefix\doi{https://doi.org/10.1016/j.ecoinf.2022.101790}.
\bibitem[{Li et~al.(2022)Li, Wang, Xiong, Cai, Yan, Yang, Liu, Fan and
  Liu}]{crestereo}
\bibinfo{author}{Li, J.}, \bibinfo{author}{Wang, P.}, \bibinfo{author}{Xiong,
  P.}, \bibinfo{author}{Cai, T.}, \bibinfo{author}{Yan, Z.},
  \bibinfo{author}{Yang, L.}, \bibinfo{author}{Liu, J.}, \bibinfo{author}{Fan,
  H.}, \bibinfo{author}{Liu, S.}, \bibinfo{year}{2022}.
\newblock \bibinfo{title}{Practical stereo matching via cascaded recurrent
  network with adaptive correlation}, in: \bibinfo{booktitle}{Proceedings of
  the IEEE/CVF Conference on Computer Vision and Pattern Recognition}, pp.
  \bibinfo{pages}{16263--16272}.
\bibitem[{Lin et~al.(2014)Lin, Maire, Belongie, Bourdev, Girshick, Hays,
  Perona, Ramanan, Doll{\'{a}}r and Zitnick}]{coco}
\bibinfo{author}{Lin, T.}, \bibinfo{author}{Maire, M.},
  \bibinfo{author}{Belongie, S.J.}, \bibinfo{author}{Bourdev, L.D.},
  \bibinfo{author}{Girshick, R.B.}, \bibinfo{author}{Hays, J.},
  \bibinfo{author}{Perona, P.}, \bibinfo{author}{Ramanan, D.},
  \bibinfo{author}{Doll{\'{a}}r, P.}, \bibinfo{author}{Zitnick, C.L.},
  \bibinfo{year}{2014}.
\newblock \bibinfo{title}{Microsoft {COCO:} common objects in context}.
\newblock \bibinfo{journal}{CoRR} \bibinfo{volume}{abs/1405.0312}.
\newblock \URLprefix \url{http://arxiv.org/abs/1405.0312},
  \href{http://arxiv.org/abs/1405.0312}{{\tt arXiv:1405.0312}}.
\bibitem[{Longuet-Higgins(1981)}]{8pointalg}
\bibinfo{author}{Longuet-Higgins, H.C.}, \bibinfo{year}{1981}.
\newblock \bibinfo{title}{A computer algorithm for reconstructing a scene from
  two projections}.
\newblock \bibinfo{journal}{Nature} \bibinfo{volume}{293},
  \bibinfo{pages}{133--135}.
\bibitem[{Loshchilov and Hutter(2017)}]{adamw}
\bibinfo{author}{Loshchilov, I.}, \bibinfo{author}{Hutter, F.},
  \bibinfo{year}{2017}.
\newblock \bibinfo{title}{Fixing weight decay regularization in adam}.
\newblock \bibinfo{journal}{CoRR} \bibinfo{volume}{abs/1711.05101}.
\newblock \URLprefix \url{http://arxiv.org/abs/1711.05101},
  \href{http://arxiv.org/abs/1711.05101}{{\tt arXiv:1711.05101}}.
\bibitem[{Maye et~al.(2013)Maye, Furgale and Siegwart}]{kalibr}
\bibinfo{author}{Maye, J.}, \bibinfo{author}{Furgale, P.},
  \bibinfo{author}{Siegwart, R.}, \bibinfo{year}{2013}.
\newblock \bibinfo{title}{Self-supervised calibration for robotic systems}, in:
  \bibinfo{booktitle}{2013 IEEE Intelligent Vehicles Symposium (IV)},
  \bibinfo{organization}{IEEE}. pp. \bibinfo{pages}{473--480}.
\bibitem[{{Mike Machado}()}]{3d_pir}
\bibinfo{author}{{Mike Machado}}, .
\newblock \bibinfo{title}{Pir sensor wall mount enclosure}.
\newblock
  \bibinfo{howpublished}{\url{https://www.thingiverse.com/thing:1718985}}.
\newblock \bibinfo{note}{This work is licensed under the Creative Commons
  Attribution 4.0 International License. To view a copy of this license, visit
  \url{http://creativecommons.org/licenses/by/4.0/}.}
\bibitem[{Moeller et~al.(2018)Moeller, Lukacs and Horne}]{moeller2018three}
\bibinfo{author}{Moeller, A.K.}, \bibinfo{author}{Lukacs, P.M.},
  \bibinfo{author}{Horne, J.S.}, \bibinfo{year}{2018}.
\newblock \bibinfo{title}{Three novel methods to estimate abundance of unmarked
  animals using remote cameras}.
\newblock \bibinfo{journal}{Ecosphere} \bibinfo{volume}{9},
  \bibinfo{pages}{e02331}.
\bibitem[{Nakashima et~al.(2018)Nakashima, Fukasawa and
  Samejima}]{nakashima2018estimating}
\bibinfo{author}{Nakashima, Y.}, \bibinfo{author}{Fukasawa, K.},
  \bibinfo{author}{Samejima, H.}, \bibinfo{year}{2018}.
\newblock \bibinfo{title}{Estimating animal density without individual
  recognition using information derivable exclusively from camera traps}.
\newblock \bibinfo{journal}{Journal of Applied Ecology} \bibinfo{volume}{55},
  \bibinfo{pages}{735--744}.
\bibitem[{Newey et~al.(2015)Newey, Davidson, Nazir, Fairhurst, Verdicchio,
  Irvine and van~der Wal}]{camera_trap_limitations}
\bibinfo{author}{Newey, S.}, \bibinfo{author}{Davidson, P.},
  \bibinfo{author}{Nazir, S.}, \bibinfo{author}{Fairhurst, G.},
  \bibinfo{author}{Verdicchio, F.}, \bibinfo{author}{Irvine, R.J.},
  \bibinfo{author}{van~der Wal, R.}, \bibinfo{year}{2015}.
\newblock \bibinfo{title}{Limitations of recreational camera traps for wildlife
  management and conservation research: A practitioner’s perspective}.
\newblock \bibinfo{journal}{Ambio} \bibinfo{volume}{44},
  \bibinfo{pages}{624--635}.
\bibitem[{Olson(2011)}]{apriltag}
\bibinfo{author}{Olson, E.}, \bibinfo{year}{2011}.
\newblock \bibinfo{title}{Apriltag: A robust and flexible visual fiducial
  system}, in: \bibinfo{booktitle}{2011 IEEE international conference on
  robotics and automation}, \bibinfo{organization}{IEEE}. pp.
  \bibinfo{pages}{3400--3407}.
\bibitem[{{Pavel Stoudek}()}]{3d_bat}
\bibinfo{author}{{Pavel Stoudek}}, .
\newblock \bibinfo{title}{Tattu 4s 6750mah lipo battery}.
\newblock
  \bibinfo{howpublished}{\url{https://grabcad.com/library/tattu-4s-6750mah-lipo-battery-1}}.
\bibitem[{Ranftl et~al.(2021)Ranftl, Bochkovskiy and Koltun}]{dpt}
\bibinfo{author}{Ranftl, R.}, \bibinfo{author}{Bochkovskiy, A.},
  \bibinfo{author}{Koltun, V.}, \bibinfo{year}{2021}.
\newblock \bibinfo{title}{Vision transformers for dense prediction}.
\newblock \bibinfo{journal}{arXiv preprint arXiv:2103.13413} .
\bibitem[{Ranftl et~al.(2020)Ranftl, Lasinger, Hafner, Schindler and
  Koltun}]{midas}
\bibinfo{author}{Ranftl, R.}, \bibinfo{author}{Lasinger, K.},
  \bibinfo{author}{Hafner, D.}, \bibinfo{author}{Schindler, K.},
  \bibinfo{author}{Koltun, V.}, \bibinfo{year}{2020}.
\newblock \bibinfo{title}{Towards robust monocular depth estimation: Mixing
  datasets for zero-shot cross-dataset transfer}.
\newblock \bibinfo{journal}{IEEE Transactions on Pattern Analysis and Machine
  Intelligence (TPAMI)} .
\bibitem[{Reconyx()}]{reconyx_hp2xc}
\bibinfo{author}{Reconyx}, .
\newblock \bibinfo{title}{Hp2xc hyperfire 2 cellular professional covert ir
  camera}.
\newblock
  \bibinfo{note}{\url{https://www.reconyx.com/product/hyperfire-2-cellular-professional-covert-ir-camera}}.
\bibitem[{Rowcliffe et~al.(2008)Rowcliffe, Field, Turvey and
  Carbone}]{rowcliffe2008estimating}
\bibinfo{author}{Rowcliffe, J.M.}, \bibinfo{author}{Field, J.},
  \bibinfo{author}{Turvey, S.T.}, \bibinfo{author}{Carbone, C.},
  \bibinfo{year}{2008}.
\newblock \bibinfo{title}{Estimating animal density using camera traps without
  the need for individual recognition}.
\newblock \bibinfo{journal}{Journal of Applied Ecology} \bibinfo{volume}{45},
  \bibinfo{pages}{1228--1236}.
\bibitem[{Shrader et~al.(2006)Shrader, Ferreira and
  Van~Aarde}]{shrader2006digital}
\bibinfo{author}{Shrader, A.M.}, \bibinfo{author}{Ferreira, S.M.},
  \bibinfo{author}{Van~Aarde, R.J.}, \bibinfo{year}{2006}.
\newblock \bibinfo{title}{Digital photogrammetry and laser rangefinder
  techniques to measure african elephants}.
\newblock \bibinfo{journal}{South African Journal of Wildlife Research-24-month
  delayed open access} \bibinfo{volume}{36}, \bibinfo{pages}{1--7}.
\bibitem[{Sixdenier et~al.(2022)Sixdenier, Wildermann, Ziegler and
  Teich}]{basestation}
\bibinfo{author}{Sixdenier, P.L.}, \bibinfo{author}{Wildermann, S.},
  \bibinfo{author}{Ziegler, D.}, \bibinfo{author}{Teich, J.},
  \bibinfo{year}{2022}.
\newblock \bibinfo{title}{Sidam: A design space exploration framework for
  multi-sensor embedded systems powered by energy harvesting}, in:
  \bibinfo{editor}{Orailoglu, A.}, \bibinfo{editor}{Reichenbach, M.},
  \bibinfo{editor}{Jung, M.} (Eds.), \bibinfo{booktitle}{Embedded Computer
  Systems: Architectures, Modeling, and Simulation},
  \bibinfo{publisher}{Springer International Publishing},
  \bibinfo{address}{Cham}. pp. \bibinfo{pages}{329--345}.
\bibitem[{Sofiiuk et~al.(2021)Sofiiuk, Petrov and Konushin}]{ritm}
\bibinfo{author}{Sofiiuk, K.}, \bibinfo{author}{Petrov, I.A.},
  \bibinfo{author}{Konushin, A.}, \bibinfo{year}{2021}.
\newblock \bibinfo{title}{Reviving iterative training with mask guidance for
  interactive segmentation}.
\newblock \bibinfo{journal}{arXiv preprint arXiv:2102.06583} .
\bibitem[{{Sony Semiconductor Solutions Corporation}()}]{imx477_datasheet}
\bibinfo{author}{{Sony Semiconductor Solutions Corporation}}, .
\newblock \bibinfo{title}{Imx477-aack product information}.
\newblock
  \bibinfo{howpublished}{\url{https://www.sony-semicon.co.jp/products/common/pdf/IMX477-AACK_Flyer.pdf}}.
\bibitem[{Srivastava et~al.(2014)Srivastava, Hinton, Krizhevsky, Sutskever and
  Salakhutdinov}]{dropout}
\bibinfo{author}{Srivastava, N.}, \bibinfo{author}{Hinton, G.},
  \bibinfo{author}{Krizhevsky, A.}, \bibinfo{author}{Sutskever, I.},
  \bibinfo{author}{Salakhutdinov, R.}, \bibinfo{year}{2014}.
\newblock \bibinfo{title}{Dropout: a simple way to prevent neural networks from
  overfitting}.
\newblock \bibinfo{journal}{The journal of machine learning research}
  \bibinfo{volume}{15}, \bibinfo{pages}{1929--1958}.
\bibitem[{{Steven Minichiello}()}]{3d_jetson}
\bibinfo{author}{{Steven Minichiello}}, .
\newblock \bibinfo{title}{nvidia jetson nano development board}.
\newblock
  \bibinfo{howpublished}{\url{https://grabcad.com/library/nvidia-jetson-nano-development-board-1}}.
\bibitem[{Sullivan et~al.(2012)Sullivan, Ohm, Han and Wiegand}]{hevc}
\bibinfo{author}{Sullivan, G.J.}, \bibinfo{author}{Ohm, J.R.},
  \bibinfo{author}{Han, W.J.}, \bibinfo{author}{Wiegand, T.},
  \bibinfo{year}{2012}.
\newblock \bibinfo{title}{Overview of the high efficiency video coding (hevc)
  standard}.
\newblock \bibinfo{journal}{IEEE Transactions on Circuits and Systems for Video
  Technology} \bibinfo{volume}{22}, \bibinfo{pages}{1649--1668}.
\newblock \DOIprefix\doi{10.1109/TCSVT.2012.2221191}.
\bibitem[{Thomas et~al.(2010)Thomas, Buckland, Rexstad, Laake, Strindberg,
  Hedley, Bishop, Marques and Burnham}]{distance_software}
\bibinfo{author}{Thomas, L.}, \bibinfo{author}{Buckland, S.T.},
  \bibinfo{author}{Rexstad, E.A.}, \bibinfo{author}{Laake, J.L.},
  \bibinfo{author}{Strindberg, S.}, \bibinfo{author}{Hedley, S.L.},
  \bibinfo{author}{Bishop, J.R.}, \bibinfo{author}{Marques, T.A.},
  \bibinfo{author}{Burnham, K.P.}, \bibinfo{year}{2010}.
\newblock \bibinfo{title}{Distance software: design and analysis of distance
  sampling surveys for estimating population size}.
\newblock \bibinfo{journal}{Journal of Applied Ecology} \bibinfo{volume}{47},
  \bibinfo{pages}{5--14}.
\newblock \URLprefix
  \url{https://besjournals.onlinelibrary.wiley.com/doi/abs/10.1111/j.1365-2664.2009.01737.x},
  \DOIprefix\doi{https://doi.org/10.1111/j.1365-2664.2009.01737.x},
  \href{http://arxiv.org/abs/https://besjournals.onlinelibrary.wiley.com/doi/pdf/10.1111/j.1365-2664.2009.01737.x}{{\tt
  arXiv:https://besjournals.onlinelibrary.wiley.com/doi/pdf/10.1111/j.1365-2664.2009.01737.x}}.
\bibitem[{Vandewalle and Varekamp(2014)}]{disparity_map_quality}
\bibinfo{author}{Vandewalle, P.}, \bibinfo{author}{Varekamp, C.},
  \bibinfo{year}{2014}.
\newblock \bibinfo{title}{Disparity map quality for image-based rendering based
  on multiple metrics}, in: \bibinfo{booktitle}{2014 International Conference
  on 3D Imaging (IC3D)}, \bibinfo{organization}{IEEE}. pp.
  \bibinfo{pages}{1--5}.
\bibitem[{Wägele et~al.(2022)Wägele, Bodesheim, Bourlat, Denzler,
  Diepenbroek, Fonseca, Frommolt, Geiger, Gemeinholzer, Glöckner, Haucke,
  Kirse, Kölpin, Kostadinov, Kühl, Kurth, Lasseck, Liedke, Losch, Müller,
  Petrovskaya, Piotrowski, Radig, Scherber, Schoppmann, Schulz, Steinhage,
  Tschan, Vautz, Velotto, Weigend and Wildermann}]{waegele_ammod}
\bibinfo{author}{Wägele, J.}, \bibinfo{author}{Bodesheim, P.},
  \bibinfo{author}{Bourlat, S.J.}, \bibinfo{author}{Denzler, J.},
  \bibinfo{author}{Diepenbroek, M.}, \bibinfo{author}{Fonseca, V.},
  \bibinfo{author}{Frommolt, K.H.}, \bibinfo{author}{Geiger, M.F.},
  \bibinfo{author}{Gemeinholzer, B.}, \bibinfo{author}{Glöckner, F.O.},
  \bibinfo{author}{Haucke, T.}, \bibinfo{author}{Kirse, A.},
  \bibinfo{author}{Kölpin, A.}, \bibinfo{author}{Kostadinov, I.},
  \bibinfo{author}{Kühl, H.S.}, \bibinfo{author}{Kurth, F.},
  \bibinfo{author}{Lasseck, M.}, \bibinfo{author}{Liedke, S.},
  \bibinfo{author}{Losch, F.}, \bibinfo{author}{Müller, S.},
  \bibinfo{author}{Petrovskaya, N.}, \bibinfo{author}{Piotrowski, K.},
  \bibinfo{author}{Radig, B.}, \bibinfo{author}{Scherber, C.},
  \bibinfo{author}{Schoppmann, L.}, \bibinfo{author}{Schulz, J.},
  \bibinfo{author}{Steinhage, V.}, \bibinfo{author}{Tschan, G.F.},
  \bibinfo{author}{Vautz, W.}, \bibinfo{author}{Velotto, D.},
  \bibinfo{author}{Weigend, M.}, \bibinfo{author}{Wildermann, S.},
  \bibinfo{year}{2022}.
\newblock \bibinfo{title}{Towards a multisensor station for automated
  biodiversity monitoring}.
\newblock \bibinfo{journal}{Basic and Applied Ecology} \bibinfo{volume}{59},
  \bibinfo{pages}{105--138}.
\newblock \URLprefix
  \url{https://www.sciencedirect.com/science/article/pii/S1439179122000032},
  \DOIprefix\doi{https://doi.org/10.1016/j.baae.2022.01.003}.
\bibitem[{Zeiler and Fergus(2014)}]{cnnvis}
\bibinfo{author}{Zeiler, M.D.}, \bibinfo{author}{Fergus, R.},
  \bibinfo{year}{2014}.
\newblock \bibinfo{title}{Visualizing and understanding convolutional
  networks}, in: \bibinfo{booktitle}{European conference on computer vision},
  \bibinfo{organization}{Springer}. pp. \bibinfo{pages}{818--833}.

\end{thebibliography}

\end{document}